\pdfoutput=1
\documentclass{article}

\PassOptionsToPackage{numbers,compress}{natbib}
\usepackage[preprint]{neurips_2026}

\usepackage[utf8]{inputenc}
\usepackage[T1]{fontenc}
\usepackage{hyperref}
\usepackage{url}
\usepackage{booktabs}
\usepackage{tabularx}
\usepackage{array}
\usepackage{multirow}
\usepackage{amsmath}
\usepackage{amssymb}
\usepackage{nicefrac}
\usepackage{microtype}
\usepackage{xcolor}
\usepackage{graphicx}
\usepackage{enumitem}
\usepackage{pifont}
\usepackage{placeins}
\usepackage{algorithm}
\usepackage{algorithmic}

\usepackage{titlesec}
\titlespacing*{\section}{0pt}{0.8ex plus 0.25ex minus 0.1ex}{0.55ex plus 0.15ex minus 0.1ex}
\titlespacing*{\subsection}{0pt}{0.75ex plus 0.25ex minus 0.1ex}{0.35ex plus 0.1ex}
\titlespacing*{\paragraph}{0pt}{0.35ex plus 0.2ex minus 0.1ex}{0.8em}

\newcommand{\cmark}{\ding{51}}
\newcommand{\xmark}{\ding{55}}

\newcommand{\ppaint}{\textsc{PPaint}}
\newcommand{\ppdistill}{\textsc{PSDistill}}
\newcommand{\srcc}{\mathrm{SRCC}}
\newcommand{\plcc}{\mathrm{PLCC}}
\newcommand{\pwdistill}{PW-Distill}
\newcommand{\dbtl}{Davidson Bradley--Terry}
\newcommand{\kendallw}{Kendall's $W$}
\newcommand{\fleissk}{Fleiss' $\kappa$}
\newcommand{\kalpha}{Krippendorff's $\alpha$}

\title{Preferences Order, Ratings Anchor: From Fused Expert Aesthetic Ground Truth to Self-Distillation}

\author{%
\begin{tabular}{c}
Yuanpei Zhao\textsuperscript{1,2,*,\textdagger}\quad
Jie Lin\textsuperscript{2,*}\quad
Chao Zhang\textsuperscript{2}\quad
Yilin Wang\textsuperscript{3}\\
Mao Li\textsuperscript{1}\quad
Chenhui Li\textsuperscript{3}\quad
Jie Hou\textsuperscript{2}\quad
Tangjie Lv\textsuperscript{2}\\[0.4em]
{\normalfont
\textsuperscript{1}Sichuan University \quad
\textsuperscript{2}NetEase Fuxi AI Lab \quad
\textsuperscript{3}East China Normal University}
\end{tabular}
}

\begin{document}

\maketitle

\begingroup
\renewcommand{\thefootnote}{\fnsymbol{footnote}}
\footnotetext[1]{Equal contribution.}
\footnotetext[2]{Work done during an internship at NetEase Fuxi AI Lab.}
\endgroup

\begin{abstract}
Pairwise preferences and pointwise ratings are the two dominant annotation protocols in image aesthetic assessment (IAA), yet existing benchmarks adopt only one, leaving their complementarity unmeasured under controlled conditions. We introduce \ppaint{}, a matched dual-protocol benchmark in which 15 domain experts (5 per category) annotate 150 Chinese paintings under both protocols across five aesthetic dimensions, collecting 45{,}900 pairwise expert judgments through a locally dense preference design alongside the matched ratings. The matched design reveals complementary strengths: preferences yield more consistent ordinal rankings, while ratings anchor the absolute score scale. Fusing both signals via two independent preference-to-score methods yields a fused expert ground truth on which the two constructions converge to nearly identical scores. The same preference-to-score principle extends to label-free VLM training. \ppdistill{} converts VLM pairwise judgments into calibrated pseudo-scores via an Elo reference pool, and trains the same VLM with confidence-weighted ranking optimization to produce a single-pass aesthetic scorer. Trained on a single painting category, the distilled Qwen3-VL-8B improves mean \(\srcc\) from 0.504 to 0.709 across all three categories, outperforming all open-source baselines including the dedicated aesthetic model ArtiMuse and matching closed-source Gemini-3.1-Pro within 0.04 \(\srcc\) at single-pass inference cost, with cross-domain transfer further validated on APDDv2.\ We will release the full \ppaint{} dataset and training code.

\end{abstract}

\begin{figure}[!ht]
\centering
\includegraphics[width=0.84\linewidth]{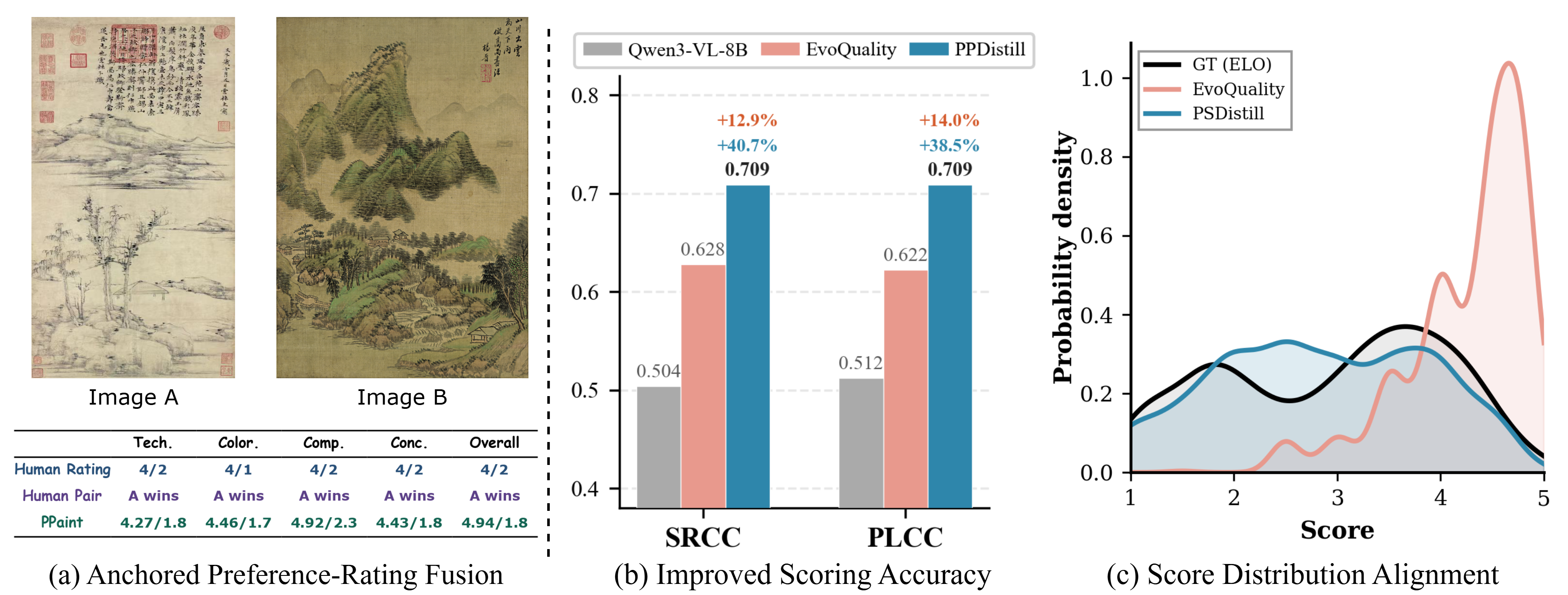}
\caption{\textbf{\ppaint{} fuses expert preferences and ratings into a calibrated fused expert ground truth for Chinese-painting aesthetics, and \ppdistill{} applies the same fusion principle to VLM pseudo-label construction, producing accurate single-pass scores.} \textbf{(a)}~\ppaint{} combines reliable rating anchors with pairwise preferences to produce fine-grained scores for each aesthetic dimension. \textbf{(b)}~On \ppaint{}, \ppdistill{} substantially improves over the base Qwen3-VL-8B and EvoQuality in mean \(\srcc\) and \(\plcc\). \textbf{(c)}~\ppdistill{} tracks the fused expert ground-truth distribution, while EvoQuality collapses to a narrow high-score band.}
\label{fig:firstfig}
\end{figure}

\section{Introduction}
\label{sec:introduction}

Aesthetic assessment relies on ground-truth labels that faithfully reflect human judgment. Expert judgments are typically elicited through one of two annotation protocols: absolute \emph{ratings} on a numerical scale (e.g., 1--5) or relative \emph{preferences} between pairs of images (A wins/Tie/B wins). The dominant practice constructs ground truth by averaging ratings across annotators~\citep{murray2012ava,kong2016aadb,wu2024qalign}, but because annotators calibrate the score range differently, a rating average conflates quality variation with inter-annotator scale usage~\citep{perezortiz2019unified,tripathi2025pairwise_pointwise}. Pairwise preferences mitigate this calibration problem by asking only which of two items is better~\citep{chiang2024chatbot_arena,li2026genarena}, yet without additional scaling or anchors, preferences do not produce the interpretable scores that downstream models require. Neither protocol alone yields an evaluation target that is both reliably ordered and grounded in an interpretable score scale.

Chinese painting provides a focused testbed for studying how annotation protocol shapes ground truth because expert evaluation follows explicit, domain-specific aesthetic criteria and remains inherently subjective~\citep{zhang2024tcp,yang2026hanmovlm}. Existing resources do not provide matched ratings and preferences with reliability diagnostics, leaving protocol effects entangled with annotator and content differences (Table~\ref{tab:gt_requirements}).

\begin{table}[!htbp]
\centering
\caption{Prior resources for aesthetic ground-truth construction. Multi-dim.\ = multi-dimensional annotation. Reliability = per-annotator data or agreement metrics available for independent assessment. Local--global structure indicates whether the resource supports both local relative comparisons and a global interpretable score scale.}
\label{tab:gt_requirements}
\scriptsize
\setlength{\tabcolsep}{3pt}
\renewcommand{\arraystretch}{1.0}
\begin{tabularx}{\linewidth}{@{}>{\raggedright\arraybackslash}p{0.22\linewidth}*{5}{>{\centering\arraybackslash}X}@{}}
\toprule
\textbf{Resource} & \textbf{Multi-dim.} & \textbf{Pointwise} & \textbf{Pairwise} & \textbf{Reliability} & \textbf{Local--global} \\
\midrule
AADB~\citep{kong2016aadb} & \cmark & \cmark & \xmark & \cmark & \xmark \\
BAID~\citep{yi2023baid} & \xmark & \cmark & \xmark & \xmark & \xmark \\
APDDv2~\citep{lu2024apddv2} & \cmark & \cmark & \xmark & \xmark & \xmark \\
ArtiMuse~\citep{cao2026artimuse} & \cmark & \cmark & \xmark & \xmark & \xmark \\
FGAesthetics~\citep{yang2026fgaesthetics} & \xmark & \xmark & \cmark & \xmark & \xmark \\
HanMo-Bench~\citep{yang2026hanmovlm} & \cmark & \cmark & \xmark & \xmark & \xmark \\
\addlinespace[1pt]
\textbf{\ppaint{} (ours)} & \cmark & \cmark & \cmark & \cmark & \cmark \\
\bottomrule
\end{tabularx}
\end{table}

The comparison in Table~\ref{tab:gt_requirements} highlights two limitations for aesthetic ground-truth construction. First, without matched ratings and preferences on the same images from the same experts, protocol effects cannot be separated from content or annotator effects. Second, without local--global structure, evaluation targets either provide an interpretable score scale without direct local preference evidence, or provide local relative preferences without a calibrated global scale. To address these limitations, we construct \ppaint{}, a dense expert-curated diagnostic benchmark in which 15 domain experts (five per category) annotate 150 Chinese paintings from three categories (flower-and-bird, landscape, and figure painting) across five aesthetic dimensions under both rating and preference protocols. The matched design reveals that preferences and ratings carry complementary information: pairwise preferences yield higher inter-expert agreement (\kendallw{}: 0.795 vs.\ 0.697), while high-agreement ratings serve as reliable anchors for the 1--5 scale, and combining the two yields a fused expert ground truth on which two independent fusion procedures converge to nearly identical scores (Figure~\ref{fig:firstfig}a). The resulting fused expert ground truth turns this complementarity into a calibrated evaluation target for testing whether models preserve both aesthetic order and score-scale alignment.

Because high-quality expert labels are expensive at training scale, we study label-free self-distillation as a cost-effective path for training on the unlabeled corpus. VLM-generated supervision faces the same preference-to-score issue: raw pointwise scores are poorly calibrated, and majority-vote preferences improve ranking but collapse into a narrow score band (Figure~\ref{fig:firstfig}c). Following this preference-to-score principle, \ppdistill{} converts VLM preferences into calibrated pseudo-scores via a small Elo reference pool and trains the same VLM with confidence-weighted GRPO into a single-pass five-dimensional scorer. Trained on one painting category, the distilled Qwen3-VL-8B improves mean \(\srcc\) from 0.504 to 0.709 across all three categories (Figure~\ref{fig:firstfig}b), achieving the best open-source pointwise result. It also comes within 0.04 \(\srcc\) of Gemini-3.1-Pro, our top closed-source pointwise baseline, and transfers to APDDv2 oil-painting and sketching subsets~\citep{lu2024apddv2}.

Our contributions can be summarized as follows:
\begin{itemize}[leftmargin=*,itemsep=1pt,topsep=1pt]
    \item \textbf{Anchored preference-rating fusion.} We propose an anchored fusion that combines preference-derived rankings with rating-based scale calibration, producing evaluation targets with both stable orderings and interpretable scores.
    \item \textbf{Matched dual-protocol benchmark.} \ppaint{} provides 150 expert-annotated paintings under matched rating and preference protocols (5 dimensions, 5 experts per category), with a high-density 612-image-pair comparison set per category that preserves local comparison information for fine-grained ranking. It also includes approximately 9{,}000 unlabeled training images for pseudo-label construction, and we will release raw annotations, fused expert ground truth, and code.
\item \textbf{Preference-to-score self-distillation.} \ppdistill{} applies the same fusion principle to VLM pseudo-label construction. An Elo-based reference pool converts VLM preferences into calibrated pseudo-scores, and confidence-weighted GRPO trains the same VLM into a single-pass five-dimensional scorer, yielding both higher ranking accuracy and better calibrated score distributions than pairwise majority-vote distillation.
\end{itemize}

\section{Related Work}
\label{sec:related_work}

\paragraph{Aesthetic assessment and Chinese painting benchmarks.}
Early IAA work established rating-based supervision through datasets such as AVA~\citep{murray2012ava} and AADB~\citep{kong2016aadb} and models such as NIMA~\citep{talebi2018nima}. Recent VLM-based scorers continue to treat aesthetic assessment primarily as pointwise score prediction~\citep{wu2024qalign,ke2023vila,zhou2024uniaa}. Art-oriented resources extend this setting with richer domains or attributes: BAID~\citep{yi2023baid} provides a large-scale artistic IAA dataset with overall aesthetic scores aggregated from online user votes, ArtiMuse~\citep{cao2026artimuse} provides eight-dimensional aesthetic scoring with expert annotations, FGAesthetics~\citep{yang2026fgaesthetics} provides pairwise comparison annotations within visually similar image series, APDDv2~\citep{lu2024apddv2} spans 24 painting and drawing categories with ten aesthetic attribute scores per image, and HanMo-Bench~\citep{yang2026hanmovlm} targets professional Chinese-painting evaluation for VLMs. These resources broaden domain coverage and annotation detail, but none provides matched pointwise and pairwise annotations from the same experts on the same images.

\paragraph{Pairwise evaluation and quality scale construction.}
Pairwise evaluation reduces the need for annotators to share a common numerical scale. Chatbot Arena~\citep{chiang2024chatbot_arena} and GenArena~\citep{li2026genarena} demonstrate effective pairwise-preference aggregation for LLM and visual-generation evaluation, respectively, and Lee and Kim~\citep{lee2019pairwise} show that pairwise comparisons can unify score regression and classification for image aesthetics. In machine translation, \citet{song2025comparative} find that side-by-side annotation yields higher inter-annotator agreement than pointwise error marking. These studies show that pairwise rankings can reduce scale-dependent noise, while also indicating that protocol choice introduces its own biases~\citep{tripathi2025pairwise_pointwise}. \citet{bansal2024peering} further report that ratings and rankings disagree on a substantial fraction of items even when collected from the same annotators. Statistical methods for converting pairwise outcomes into global quality scales are well established~\citep{thurstone1927law,bradley1952rank,davidson1970ties,elo1978rating,perezortiz2019unified}. To our knowledge, no prior work provides a matched expert comparison within a single artistic domain that quantifies the protocol tradeoff and fuses both signals into a calibrated scale.

\paragraph{Preference-based training for visual scoring.}
Compare2Score~\citep{zhu2025compare2score} uses multimodal pairwise comparison to derive continuous image quality scores, but focuses on single-dimensional IQA rather than multi-dimensional expert-scale construction. VisualQuality-R1~\citep{chen2025visualqualityr1} trains a single-dimension IQA scorer via reinforcement learning to rank with human MOS as ground truth, and EvoQuality~\citep{chen2025evoquality} extends this framework to a label-free setting using model-generated majority-vote preferences. Our approach differs in two respects: the offline stage converts VLM pairwise preferences into multi-dimensional pseudo-scores on a shared 1--5 score scale, and the distillation uses confidence weighting to account for uneven pseudo-label reliability.

\section{\ppaint{}: Matched Protocols and Fused Expert Ground Truth}
\label{sec:cpa_gt}

\subsection{Data Collection and Annotation Protocol}
\label{sec:cpa_gt_annotation}

The dataset comprises 150 carefully curated Chinese paintings from three categories with 50 paintings per category (flower-and-bird, landscape, and figure painting), each evaluated along five aesthetic dimensions: brush technique, coloration, composition, artistic conception, and overall quality. Five domain experts per category annotate all images under both rating and preference protocols, with images, experts, and dimensions held fixed across protocols.

Under the rating protocol, experts assign an integer score from 1 to 5 to each image--dimension pair, producing 3{,}750 raw scores. Under the preference protocol, the same experts compare two paintings from the same category under a given dimension and select \textsc{A wins}, \textsc{Tie}, or \textsc{B wins}, yielding 45{,}900 pairwise judgments. Each category uses 612 of the 1{,}225 possible pairs (50\% coverage), and all five category experts annotate the same pairs. The pairwise budget determines how densely expert preferences cover local quality relations within a category. Exhaustive annotation of all 1{,}225 pairs is costly, but reducing the budget too far removes comparisons needed to distinguish nearby-quality paintings and makes the recovered ranking depend heavily on indirect inference. \ppaint{} therefore uses 612 pairs per category as a high-density half-coverage design: it retains enough local preference information to recover fine-grained rankings while keeping expert workload tractable. Consistent with this design goal, the comparison set remains tightly connected (diameter 2 in all three categories), and a proxy subsampling study shows that Elo rankings at this budget reach 0.986 \(\srcc\) and 0.989 \(\plcc\) against the full 1{,}225-pair reference (Appendix~\ref{app:pair_budget}). Full annotation statistics, protocol order, interface details, expert qualifications, and quality-control procedures are reported in Appendix~\ref{app:dataset}.

\begin{figure}[t]
\centering
\includegraphics[width=0.95\linewidth]{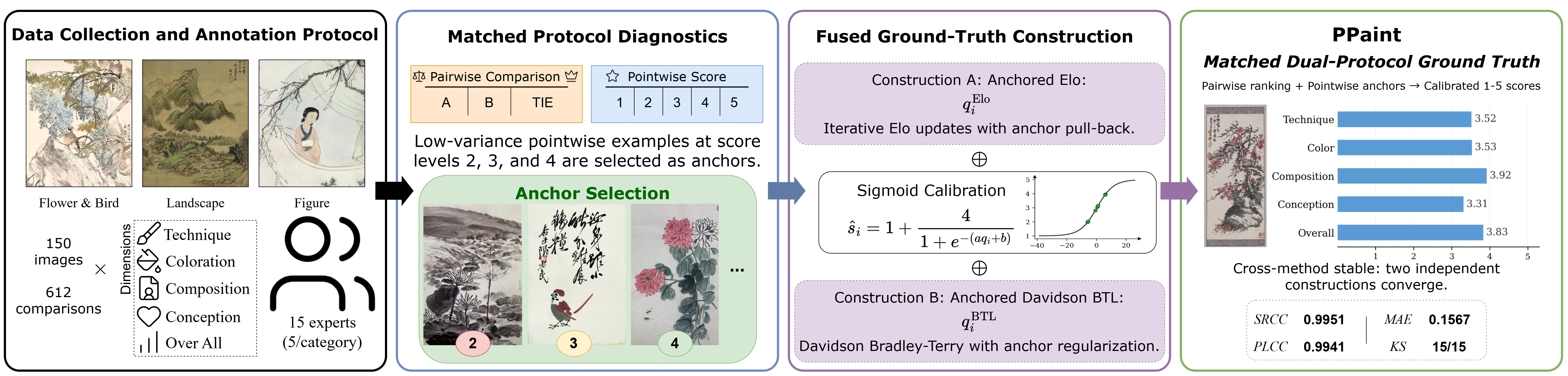}
\caption{Overview of the preference-rating fusion for the fused expert ground truth. Five category-specific experts annotate the same 150 paintings under both rating (1--5 scores) and preference (\textsc{A wins}/\textsc{Tie}/\textsc{B wins}) protocols across five aesthetic dimensions. Low-variance rating examples serve as anchors for the expert-defined 1--5 score scale. Preference judgments are aggregated into latent quality estimates by two independent methods (anchored Elo and anchored \dbtl{}), each constrained by the same anchors and mapped to a five-point scale through sigmoid calibration. The two constructions converge to nearly identical scores.}
\label{fig:cpa_gt_overview}
\end{figure}

\subsection{Matched Protocol Diagnostics}
\label{sec:cpa_gt_protocol_findings}

The matched design allows us to quantify the extent to which pairwise preferences mitigate inter-annotator scale-use differences under fixed images, experts, and dimensions. We compare the two protocols using four diagnostics (\kendallw{}, split-half Spearman correlation, weak-label triplet separation, and to-consensus pairwise ranking accuracy) and report their averages across all 15 category--dimension groups. For \kendallw{}, each category--dimension group is represented as five expert-specific rankings over the same 50 paintings: rating rankings are induced directly from each expert's 1--5 scores, whereas preference rankings are obtained by aggregating that expert's pairwise judgments with Elo. Thus, the statistic compares protocol-induced expert rankings with identical item and rater dimensions. The aggregation step is part of how pairwise annotations define a global ranking rather than an additional post hoc correction. Split-half Spearman repeatedly splits the five experts into two halves and correlates the resulting half-group rankings, quantifying how stable the consensus ranking is under expert subsampling. Triplet separation uses weak quality tiers (high, medium, low) elicited from expert discussion on the overall dimension and measures how often the protocol-induced ranking correctly separates triplets sampled across these three tiers. To-consensus PRA builds a group consensus ranking from all five experts and then averages, across experts, the pairwise ranking accuracy of each expert's protocol-induced ranking against this consensus.

\begin{table}[!htbp]
\centering
\caption{Protocol diagnostics under matched annotation (averages over 15 category--dimension groups). Higher values indicate stronger inter-expert agreement or ranking stability.}
\label{tab:protocol_diagnostics}
\scriptsize
\setlength{\tabcolsep}{3pt}
\renewcommand{\arraystretch}{1.0}
\begin{tabular}{@{}lccc@{}}
\toprule
\textbf{Diagnostic} & \textbf{Preferences} & \textbf{Ratings} & \textbf{Relative gap} \\
\midrule
\kendallw{} (expert consensus) & 0.795 & 0.697 & \(+\)14.1\% \\
Split-half Spearman (ranking stability) & 0.836 & 0.757 & \(+\)10.4\% \\
Triplet separation (weak-label ordering) & 0.92 & 0.86 & \(+\)7.0\% \\
To-consensus PRA (annotator--group agreement) & 0.787 & 0.694 & \(+\)13.4\% \\
\bottomrule
\end{tabular}
\end{table}
\FloatBarrier

Table~\ref{tab:protocol_diagnostics} shows that preferences yield higher values than ratings on all four diagnostics. The advantage is not concentrated in a single category or dimension: the \kendallw{} gain is positive in 14 of 15 category--dimension groups (Appendix~\ref{app:protocol_viz}, Figure~\ref{fig:protocol_diagnostics}), with only a negligible reversal in figure-painting brush technique. The gap tends to be larger on more subjective dimensions such as artistic conception, where the pairwise format yields the largest gains over numerical scoring. Additional control experiments that equalize the per-protocol judgment count and the per-protocol annotator count further confirm that this advantage is intrinsic to the protocol rather than a sampling artifact (Appendices~\ref{app:per_judgment} and~\ref{app:annotator_stability}).

However, each preference judgment is inherently \emph{local} and yields a latent ranking with arbitrary units that offers no cross-category comparability. Ratings are noisier in aggregate, but high-agreement images provide reliable anchors for the 1--5 rubric, supplying the global scale reference that preferences alone cannot. This complementarity motivates the fusion strategy in \S\ref{sec:cpa_gt_fusion}: preferences determine rank order, and high-agreement ratings anchor the score scale.

\subsection{Fused Expert Ground-Truth Construction}
\label{sec:cpa_gt_fusion}

We therefore construct the fused expert ground truth by combining locally dense pairwise ranking evidence with rating-based scale anchors (Figure~\ref{fig:cpa_gt_overview}). The 612-pair preference budget supplies the local information needed for fine-grained ordering, while high-agreement ratings align this order to the expert-defined 1--5 scale. Anchors are low-disagreement rating examples that span distinct score levels, so they provide reliable references on the expert-defined score scale while the preference data determines relative order. The anchors align the preference-derived latent ranking with the expert-defined 1--5 rubric. They define an expert-agreed score scale rather than a universal aesthetic scale. We instantiate this idea with two independent constructions: anchored Elo as the canonical construction and anchored \dbtl{} as a cross-method check.

All constructions share the same anchor set and final calibration. Let \(\mathcal{A}\) denote the anchor set and \(\bar{s}_i\) the mean expert rating for anchor image \(i\). The calibrated aesthetic score is then obtained by a shared sigmoid mapping,
\begin{equation}
\hat{s}_i = 1 + 4 \cdot \sigma(aq_i+b), \label{eq:sigmoid_cal}
\end{equation}
where \(a\) and \(b\) are fitted to anchors pooled across all category--dimension groups so that each group is placed on a common scale. Anchor selection details are given in Appendix~\ref{app:anchors}.

\textbf{Construction A: Anchored Elo.}\quad
We estimate latent qualities through iterative Elo updates~\citep{elo1978rating}. After each full pass over the observed pairs, anchor images are regularized toward their latent-space targets:
\begin{equation}
q_i \leftarrow q_i + \alpha(\bar{q}_i - q_i), \quad i \in \mathcal{A},
\label{eq:anchor_reg}
\end{equation}
where \(\bar{q}_i\) is the level-mapped Elo target corresponding to \(\bar{s}_i\) and $\alpha$ controls anchor strength. The calibrated score is then obtained by Eq.~\eqref{eq:sigmoid_cal}. Standard Elo update equations and hyperparameter ablations are in Appendix~\ref{app:anchored_elo} and \ref{app:elo_ablation}.

\textbf{Construction B: Anchored \dbtl{}.}\quad
As an independent check, we also fit a Davidson extension of Bradley--Terry that explicitly models \textsc{Tie} outcomes and uses the same anchor set and sigmoid calibration. Its likelihood and anchor regularizer are given in Appendix~\ref{app:anchored_btl}.

Anchored Elo and anchored \dbtl{} produce nearly identical rank orders (\(\srcc=0.9951\)), calibrated scores (\(\plcc=0.9941\)), and induced pairwise decisions (98.5\% agreement). The agreement between two mathematically distinct ranking procedures under the same anchor constraints suggests that the fused scores are not sensitive to the particular ranking estimator. Comparing anchored and unanchored variants further shows that rating-based anchors are the key factor driving convergence. Without anchor regularization, the two estimators diverge substantially. MAE on the 1--5 scale increases from 0.157 to 0.282, and only 8 of 15 category--dimension groups pass the Kolmogorov--Smirnov test, compared with 15 of 15 with anchors. See Appendix~\ref{app:cross_method}, Table~\ref{tab:four_method}. We adopt the anchored Elo output as the canonical fused expert ground truth for subsequent experiments, with additional annotator-stability and cross-method evidence reported in Appendix~\ref{app:annotation_diagnostics} and \ref{app:gt_construction}.

\section{\ppdistill{}: Preference-to-Score Self-Distillation}
\label{sec:vlm_diagnosis}

Pseudo-labels for the unlabeled training corpus must come from the model itself. Following the preference-to-score principle of \S\ref{sec:cpa_gt}, we construct \ppdistill{} in two stages (Figure~\ref{fig:method_overview}): an offline Elo-based reference pool converts VLM preferences into calibrated pseudo-scores (\S\ref{sec:bridge}), and online confidence-weighted GRPO trains the same VLM into a single-pass scorer (\S\ref{sec:grpo_method}). Compared with raw pointwise distillation (0.557 mean \(\srcc\), Table~\ref{tab:main_results}) or majority-vote preference distillation (0.628), this design directly addresses the score-scale collapse identified in Figure~\ref{fig:firstfig}c.

\subsection{Preference-to-Score Bridge via Elo Reference Pool}
\label{sec:bridge}

We adopt a reference-pool strategy inspired by Compare2Score~\citep{zhu2025compare2score}, but estimate the pool scores from the VLM's own pairwise judgments rather than from human-rated reference images.

\begin{figure}[t]
\centering
\includegraphics[width=1\linewidth]{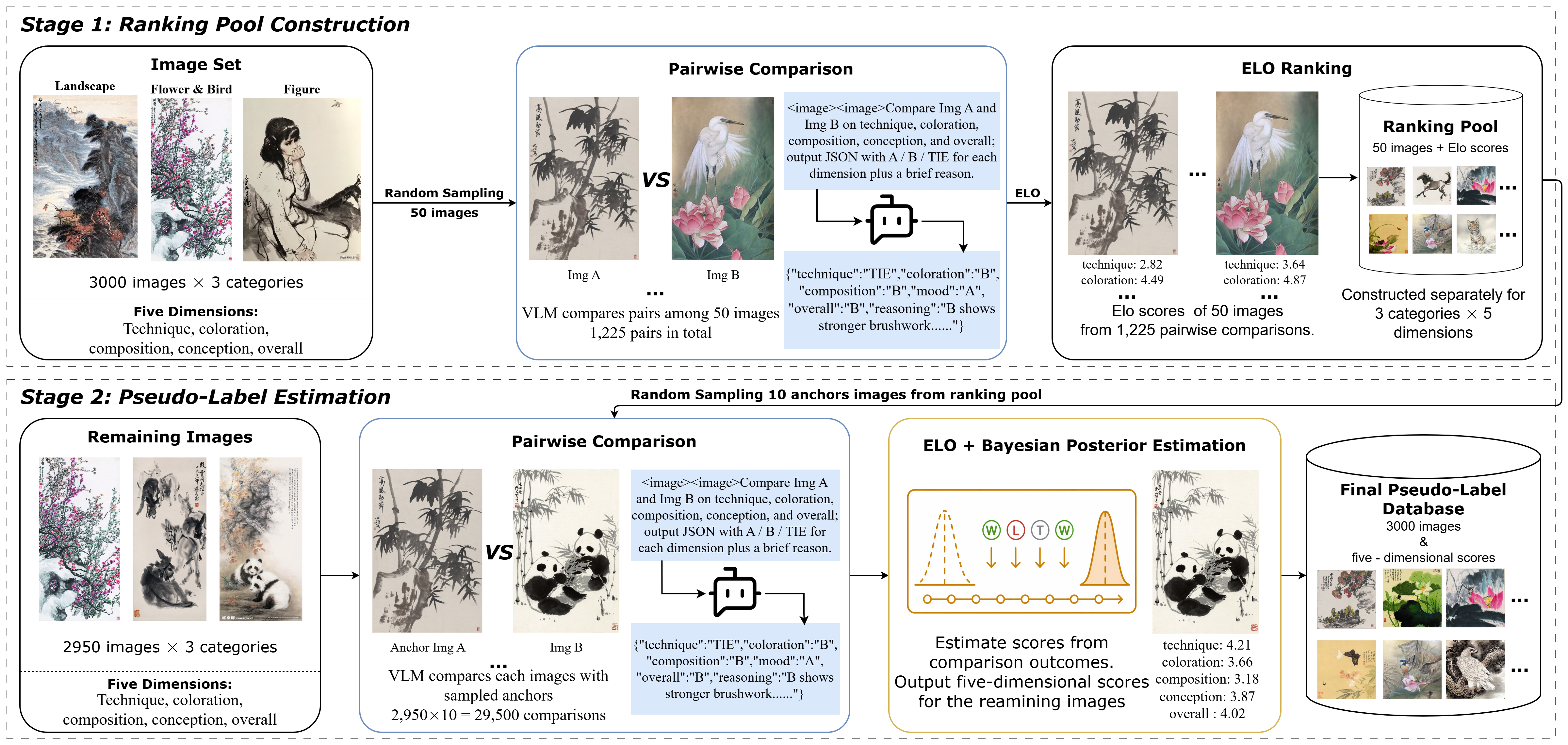}
\caption{The offline preference-to-score bridge of \ppdistill{}. \textbf{Stage~1} constructs a ranking pool via full pairwise comparison within $N$ randomly sampled images, and Elo updates produce five-dimensional reference ratings for the pool images. \textbf{Stage~2} estimates pseudo-scores for each remaining image by comparing it against $K$ pool references via Bayesian posterior inference and sigmoid calibration. The online distillation stage is described in \S\ref{sec:grpo_method}.}
\label{fig:method_overview}
\end{figure}

We first sample $N{=}50$ unlabeled training images and perform exhaustive pairwise comparison within this pool, yielding $\binom{50}{2}{=}1{,}225$ pairs (Figure~\ref{fig:method_overview}, Stage~1). Standard Elo updates~\citep{elo1978rating} produce a five-dimensional rating for every pool image (Appendix~\ref{app:anchored_elo}). The pool ratings are then frozen and serve as reference scores for placing new images on the same Elo-derived scale, playing a role analogous to the rating-derived anchors in \S\ref{sec:cpa_gt_fusion} (cf.\ Eq.~\ref{eq:anchor_reg}). Each remaining image is compared with only $K{=}10$ randomly sampled pool references, and we estimate its Elo rating from these sparse comparisons using Bayesian posterior inference.

Finally, we map raw Elo ratings to the 1--5 aesthetic scale by sigmoid calibration:
\begin{equation}
\tilde{s}_i^{(d)}=1+\frac{4}{1+\exp\!\left(-\left(\lambda
\frac{q_i^{(d)}-q_{\min}^{(d)}}{q_{\max}^{(d)}-q_{\min}^{(d)}}-\lambda/2\right)\right)},
\quad \lambda=6 .
\label{eq:bridge_calibration}
\end{equation}
where $q_i^{(d)}$ is the raw Elo rating for image $i$ on dimension $d$, and $q_{\min}^{(d)}$, $q_{\max}^{(d)}$ are the extrema across the training set. The factor 4 sets the span of the 1--5 output scale. We set $\lambda=6$ so that the normalized Elo range is mapped to the sigmoid input interval \([-3,3]\), which keeps the extrema away from the exact score boundaries. The bridge mirrors the \S\ref{sec:cpa_gt} principle: Elo supplies a global ordering from pairwise preferences and sigmoid calibration maps the Elo ratings to the 1--5 scale, mitigating the distribution collapse of Figure~\ref{fig:firstfig}c. It requires $O(N^2{+}nK)$ calls (approximately $31$K versus $4.5$M for exhaustive comparison), with the complete procedure in Algorithm~\ref{alg:bridge} (Appendix~\ref{app:pseudo_label_algo}).

\subsection{Confidence-Weighted Self-Distillation via Fidelity Reward}
\label{sec:grpo_method}

The calibrated pseudo-scores provide a usable scale, but their point estimates remain noisy because each image is observed through only $K$ reference comparisons. Rather than regress to these point estimates directly, we induce relative orderings and optimize a ranking-fidelity reward that preserves the more reliable ordinal signal while reducing sensitivity to calibration noise.

Motivated by the ranking-fidelity objective of VisualQuality-R1~\citep{chen2025visualqualityr1}, we train the model to match target orderings via GRPO~\citep{shao2024deepseekmath}. The full objective is given in Appendix~\ref{app:grpo}.

Each candidate response produces a $D{=}5$ dimensional score vector. For a minibatch pair $(i,z)$, we estimate the probability that sample $k$ for image $i$ ranks above image $z$ under a Thurstone model:
\begin{equation}
p_{\text{pred}}^{(d)} = \Phi\!\left(\frac{\hat{s}_{i,k}^{(d)} - \bar{s}_z^{(d)}}{\sqrt{\sigma_i^{2(d)} + \sigma_z^{2(d)} + \epsilon}}\right), \label{eq:thurstone}
\end{equation}
where $\Phi$ is the standard normal CDF. The target probability is defined from the ordering of the bridge pseudo-scores:
\begin{equation}
p_{\text{gt}}^{(d)} = \begin{cases} 1 & \tilde{s}_i^{(d)} > \tilde{s}_z^{(d)}, \\ 0 & \tilde{s}_i^{(d)} < \tilde{s}_z^{(d)}, \\ 0.5 & \tilde{s}_i^{(d)} = \tilde{s}_z^{(d)}. \end{cases} \label{eq:pgt_hard}
\end{equation}
We measure agreement using the Bhattacharyya coefficient~\citep{tsai2007frank}:
\begin{equation}
R_{\text{fidelity}}^{(d)} = \sqrt{p_{\text{pred}}^{(d)} \cdot p_{\text{gt}}^{(d)}} + \sqrt{(1-p_{\text{pred}}^{(d)})(1-p_{\text{gt}}^{(d)})}. \label{eq:fidelity}
\end{equation}
To reduce the influence of near-tie pairs, which are more sensitive to pseudo-label noise, we weight each pair by its calibrated score gap:
\begin{equation}
w_{iz}^{(d)} = \min\!\left(\frac{|\tilde{s}_i^{(d)} - \tilde{s}_z^{(d)}|}{\tau_w},\; 1\right), \label{eq:pair_weight}
\end{equation}
and aggregate the weighted fidelity across comparison partners and dimensions:
\begin{equation}
R_{\text{rank}}^{(i,k)} = \frac{1}{D}\sum_{d=1}^{D} \frac{\sum_{z\neq i} w_{iz}^{(d)} \cdot R_{\text{fidelity},iz}^{(d)}}{\sum_{z\neq i} w_{iz}^{(d)}}. \label{eq:weighted_reward}
\end{equation}
The resulting reward is used as the task reward in GRPO. Pairwise comparisons are used only for pseudo-label construction, and the trained model requires only a single forward pass per image at inference time.

\section{Experiments}
\label{sec:experiments}

\subsection{Experimental Setup}
\label{sec:exp_setup}

\paragraph{Datasets and evaluation.}
We train on the unlabeled flower-and-bird split of \ppaint{} (${\sim}3{,}000$ images), without using any human annotations. We use this single-category split as the default training source so that performance on landscape and figure painting directly tests cross-category generalization, and report training-source ablations in Appendix~\ref{app:cross_category}. Pseudo-labels are generated once in the offline stage (\S\ref{sec:bridge}) using the pretrained model without any parameter updates. Evaluation is conducted zero-shot on the full \ppaint{} expert-annotated test set of 150 images across three categories and five aesthetic dimensions. MD5 hashing confirms no image duplicates between the training corpus and the 150-image test set. We report \(\srcc\) and \(\plcc\) against the fused expert ground truth (\S\ref{sec:cpa_gt_fusion}). These metrics jointly measure ordinal agreement and linear correlation with the calibrated 1--5 scores. All evaluations use greedy decoding with a single forward pass per image.

\paragraph{Competing methods.}
We organize competing methods into three groups. The first group contains general-purpose VLMs evaluated under direct pointwise scoring. It includes seven closed-source models, Gemini-3.1-Pro, Gemini-3-Flash, Claude-Sonnet-4.6, Doubao-Seed-1.6-Vision, Qwen3-VL-Plus~\citep{bai2025qwen3vl}, Qwen3.6-Plus, and GPT-5.4-Mini, together with four open-source VLMs, Qwen3-VL-8B-Instruct (our distillation base), Qwen3-VL-32B-Instruct~\citep{bai2025qwen3vl}, InternVL3.5-8B~\citep{zhu2025internvl3}, and InternVL3.5-38B~\citep{zhu2025internvl3}. Closed-source model versions are reported in Appendix~\ref{app:closed_model_versions}. The second group covers task-oriented baselines that are closest to our setting. ArtiMuse~\citep{cao2026artimuse} is a recent state-of-the-art model for multi-dimensional image aesthetic assessment (IAA), and EvoQuality-style majority-vote preference distillation~\citep{chen2025evoquality} is a recent state-of-the-art label-free self-distillation method for image quality assessment (IQA). To isolate the effect of base model and training data, we evaluate ArtiMuse using the authors' released checkpoint without further fine-tuning, and we retrain the EvoQuality-style baseline and \pwdistill{} on the same Qwen3-VL-8B backbone and \ppaint{} unlabeled flower-and-bird corpus as \ppdistill{}. Finally, as a training-free reference, we report Qwen3-VL-8B with full pairwise Elo aggregation~\citep{elo1978rating}, which requires $O(n^2)$ test-time VLM calls.

\paragraph{Implementation details.}
We adopt Qwen3-VL-8B-Instruct~\citep{bai2025qwen3vl} as the policy backbone and fine-tune it via GRPO~\citep{shao2024deepseekmath} with DeepSpeed ZeRO-3 on 8$\times$NVIDIA~A100~80GB GPUs. For each image we sample $G{=}6$ candidate responses. Optimization uses AdamW with learning rate $1{\times}10^{-6}$ and gradient clipping at $1.0$. The effective batch size is $64$ images per step. Training runs for $10$ epochs with maximum completion length $512$ tokens. Prompt templates are provided in Appendix~\ref{app:prompt}.

\subsection{Main Results}
\label{sec:main_results}

\FloatBarrier
\begin{table}[!htbp]
\centering
\caption{Aesthetic scoring on \ppaint{}. Per-category and mean \(\srcc\)/\(\plcc\) against the fused expert ground truth. $\ast$: distillation base model. $\dagger$: pairwise Elo aggregation requiring $O(n^2)$ inference. $\ddagger$: trained via the same Thurstone-model-based GRPO framework as \ppdistill{} but with different pseudo-label sources (\S\ref{sec:ablation}). HN = flower-and-bird, RW = figure painting, SS = landscape. \textbf{Bold} = best among closed-source models. \underline{Underline} = best among open-source and self-distillation models with full five-dimensional scoring.}
\label{tab:main_results}
\footnotesize
\setlength{\tabcolsep}{2.1pt}
\renewcommand{\arraystretch}{0.96}
\begin{tabular}{@{}l*{4}{r}@{\hspace{0.7em}}*{4}{r}@{}}
\toprule
& \multicolumn{4}{c}{\(\srcc \uparrow\)} & \multicolumn{4}{c}{\(\plcc \uparrow\)} \\
\cmidrule(lr){2-5} \cmidrule(lr){6-9}
\textbf{Method} & \multicolumn{1}{c}{\textbf{HN}} & \multicolumn{1}{c}{\textbf{RW}} & \multicolumn{1}{c}{\textbf{SS}} & \multicolumn{1}{c}{\textbf{Mean}} & \multicolumn{1}{c}{\textbf{HN}} & \multicolumn{1}{c}{\textbf{RW}} & \multicolumn{1}{c}{\textbf{SS}} & \multicolumn{1}{c}{\textbf{Mean}} \\
\midrule
\multicolumn{9}{l}{\textit{Closed-source VLMs (direct pointwise)}} \\
Gemini-3.1-Pro        & \textbf{0.744} & 0.688 & \textbf{0.807} & \textbf{0.746} & \textbf{0.816} & \textbf{0.774} & 0.826 & \textbf{0.805} \\
Gemini-3-Flash        & 0.734 & 0.666 & 0.801 & 0.733 & 0.779 & 0.741 & \textbf{0.834} & 0.785 \\
Claude-Sonnet-4.6            & 0.669 & \textbf{0.698} & 0.732 & 0.700 & 0.710 & 0.711 & 0.720 & 0.714 \\
Qwen3.6-Plus          & 0.718 & 0.566 & 0.660 & 0.648 & 0.770 & 0.627 & 0.686 & 0.694 \\
Doubao-Seed-1.6       & 0.534 & 0.631 & 0.598 & 0.587 & 0.541 & 0.633 & 0.601 & 0.591 \\
GPT-5.4-Mini          & 0.492 & 0.622 & 0.487 & 0.534 & 0.501 & 0.630 & 0.462 & 0.531 \\
Qwen3-VL-Plus         & 0.558 & 0.647 & 0.516 & 0.574 & 0.503 & 0.635 & 0.441 & 0.526 \\
\midrule
\multicolumn{9}{l}{\textit{Open-source models (direct pointwise)}} \\
Qwen3-VL-32B          & 0.644 & \underline{0.734} & 0.610 & 0.663 & 0.660 & \underline{0.714} & 0.619 & 0.664 \\
ArtiMuse              & \underline{0.797} & 0.587 & 0.675 & 0.686 & \underline{0.758} & 0.564 & 0.592 & 0.638 \\
Qwen3-VL-8B$^\ast$    & 0.492 & 0.568 & 0.453 & 0.504 & 0.485 & 0.580 & 0.472 & 0.512 \\
InternVL3.5-38B       & 0.322 & 0.555 & 0.233 & 0.370 & 0.314 & 0.527 & 0.248 & 0.363 \\
InternVL3.5-8B        & 0.259 & 0.176 & 0.329 & 0.255 & 0.249 & 0.171 & 0.288 & 0.236 \\
\midrule
\multicolumn{9}{l}{\textit{Pairwise aggregation (training-free, $O(n^2)$ inference)}} \\
Qwen3-VL-8B$^\dagger$   & 0.665 & 0.641 & 0.609 & 0.638 & 0.692 & 0.617 & 0.583 & 0.630 \\
\midrule
\multicolumn{9}{l}{\textit{Self-distillation (single-pass inference)}} \\
EvoQuality$^\ddagger$     & 0.644 & 0.692 & 0.549 & 0.628 & 0.643 & 0.680 & 0.542 & 0.622 \\
\pwdistill{}$^\ddagger$  & 0.523 & 0.689 & 0.459 & 0.557 & 0.526 & 0.683 & 0.459 & 0.556 \\
\ppdistill{} (ours) & 0.727 & 0.713 & \underline{0.686} & \underline{0.709} & 0.743 & 0.693 & \underline{0.692} & \underline{0.709} \\
\bottomrule
\end{tabular}
\end{table}

\paragraph{Comparison to visual scoring baselines.}
Table~\ref{tab:main_results} reports per-category and mean \(\srcc\)/\(\plcc\) for all methods on \ppaint{}. \ppdistill{} reaches a mean \(\srcc\) of $0.709$, a 40.7\% relative improvement over its base and the highest score among open-source single-pass pointwise scorers with full five-dimensional outputs, surpassing the dedicated aesthetic-assessment model ArtiMuse ($0.686$) and the larger Qwen3-VL-32B ($0.663$). Although ArtiMuse performs strongly on flower-and-bird paintings, it remains below \ppdistill{} in mean \(\srcc\) ($0.686$ vs. $0.709$) and by a wider margin in mean \(\plcc\) ($0.638$ vs. $0.709$), suggesting that alignment with the target annotation protocol and score scale is important for Chinese-painting aesthetic evaluation. \ppdistill{} comes within $0.02$ \(\srcc\) of Gemini-3.1-Pro on flower-and-bird ($0.727$ vs. $0.744$) and exceeds all closed-source pointwise models on figure painting ($0.713$ vs. $0.688$ for the strongest proprietary baseline), while also surpassing the training-free pairwise Elo reference in mean \(\srcc\) ($0.638$) at single-pass inference cost. The remaining point-estimate gap to the closed-source leader is concentrated in landscape, a category absent from training data, yet landscape \(\srcc\) still rises from $0.453$ to $0.686$ (51.4\% relative improvement) and figure painting from $0.568$ to $0.713$ (25.5\% relative improvement), indicating that the distilled signal generalizes beyond the source category.

\paragraph{Effect of pseudo-label construction.}
Within the self-distillation block of Table~\ref{tab:main_results}, the three pseudo-label sources exhibit a clear ordering: Calibrated Elo attains $0.709$ mean \(\srcc\), majority vote $0.628$, and \pwdistill{} $0.557$. This ordering mirrors the protocol complementarity observed in expert annotation (\S\ref{sec:cpa_gt_protocol_findings}): majority vote preserves the ordinal direction of preferences but discards score magnitude, whereas \pwdistill{} retains a nominal scale yet inherits the calibration noise of raw pointwise outputs. Calibrated Elo combines a global preference-derived ordering with an explicit score-scale mapping, producing pseudo-labels of substantially finer granularity than either alternative (Appendix~\ref{app:pseudo_label_algo}, Table~\ref{tab:pseudo_label_stats}).

\subsection{Ablation Studies}
\label{sec:ablation}

\paragraph{Generalization beyond Chinese painting.}
To test whether the gains observed on Chinese painting carry over to substantially different visual domains, we evaluate \ppdistill{} on the oil-painting and sketching subsets of APDDv2~\citep{lu2024apddv2}. Compared with the EvoQuality baseline, \ppdistill{} delivers a $6.0\%$ to $6.6\%$ gain in mean \(\srcc\) and a $22.9\%$ to $31.7\%$ improvement over the base model (Table~\ref{tab:apddv2_results}; setup details in Appendix~\ref{app:apddv2_setup}). The advantage therefore transfers across substantially different visual statistics and rating rubrics.

\begin{table}[H]
\centering
\caption{Cross-dataset generalization on APDDv2. Best epoch per configuration is selected by mean \(\srcc\). All models start from Qwen3-VL-8B-Instruct. \textbf{Bold} marks the highest mean \(\srcc\) and mean \(\plcc\) across all configurations.}
\label{tab:apddv2_results}
\scriptsize
\setlength{\tabcolsep}{4pt}
\renewcommand{\arraystretch}{1.10}
\begin{tabular}{@{}ll*{3}{c}*{3}{c}@{}}
\toprule
& & \multicolumn{3}{c}{\(\srcc \uparrow\)} & \multicolumn{3}{c}{\(\plcc \uparrow\)} \\
\cmidrule(lr){3-5} \cmidrule(lr){6-8}
\textbf{Method} & \textbf{Train} & \textbf{Oil} & \textbf{Sketch} & \textbf{Mean} & \textbf{Oil} & \textbf{Sketch} & \textbf{Mean} \\
\midrule
Base (pretrained)     & --   & 0.540 & 0.477 & 0.508 & 0.552 & 0.495 & 0.524 \\
\midrule
EvoQuality            & Oil  & 0.683 & 0.574 & 0.628 & 0.670 & 0.575 & 0.623 \\
EvoQuality            & Skt  & 0.620 & 0.560 & 0.590 & 0.627 & 0.533 & 0.580 \\
\midrule
\ppdistill{} (ours)   & Oil  & 0.752 & 0.587 & \textbf{0.670} & 0.746 & 0.574 & \textbf{0.660} \\
\ppdistill{} (ours)   & Skt  & 0.690 & 0.559 & 0.625 & 0.678 & 0.551 & 0.615 \\
\bottomrule
\end{tabular}
\end{table}

\paragraph{Robustness to fused expert ground-truth construction.}
Beyond the fused expert ground truth above, we evaluate \ppdistill{} against pure human pointwise targets to test whether the gains hinge on the choice of target construction. The relative ordering among self-distillation variants is preserved under both targets, with mean \(\srcc\) reaching $0.621$ against the rating mean and $0.594$ against the rating median (Appendix~\ref{app:gt_robustness}, Table~\ref{tab:gt_robustness}). On the absolute 1--5 scale, the predicted scores also remain well calibrated to the fused expert ground truth, with mean absolute error approximately $40\%$ lower than that of the next-best baseline (Table~\ref{tab:gt_calibration}). The stability is consistent with the preference-based supervision in \S\ref{sec:bridge}: pairwise signals fit a global ordering rather than any particular pointwise aggregator, so changing the evaluation target only shifts the calibration constants.

\paragraph{Effect of training-source category.}
We further train both \ppdistill{} and the EvoQuality-style majority-vote baseline~\citep{chen2025evoquality} on each of the three categories separately, with full results reported in Appendix~\ref{app:cross_category} (Table~\ref{tab:cross_category}). \ppdistill{} outperforms majority vote on every per-category and mean \(\srcc\)/\(\plcc\) value regardless of training source. The spread of its mean \(\srcc\) across the three training sources is only $0.019$, less than half of the spread observed for majority vote ($0.050$) and \pwdistill{} ($0.053$). Training on figure painting yields the strongest overall mean \(\srcc\) of $0.715$, with particularly strong transfer to landscape ($0.754$ \(\srcc\)).

\paragraph{Effect of confidence weighting.}
The confidence weight in Eq.~\ref{eq:pair_weight} reduces the contribution of pairs whose calibrated score gap falls below the noise threshold. To isolate its effect, we train two variants under identical ranking-pool seed and optimization budget, with and without the weight applied. Per-category results are reported in Appendix~\ref{app:extended_analysis} (Table~\ref{tab:pair_weight}). Confidence weighting raises mean \(\srcc\) from $0.660$ to $0.709$, a relative improvement of $7.4\%$, and produces gains on all three categories. The largest gain appears on landscape ($+0.073$ \(\srcc\), $11.9\%$ relative), the category for which a larger fraction of pseudo-label pairs falls near the decision boundary.

\section{Conclusion}
\label{sec:conclusion}

Preferences yield more consistent ordinal rankings while high-agreement ratings anchor the absolute score scale, and the two signals carry complementary information for subjective aesthetic evaluation. \ppaint{} instantiates this insight as a dense matched dual-protocol benchmark for Chinese painting aesthetics, and the proposed anchored fusion produces a calibrated fused expert ground truth on which two independent preference-to-score procedures converge to nearly identical scores. \ppdistill{} extends the same principle to label-free VLM training. By converting VLM pairwise judgments into calibrated pseudo-scores via an Elo reference pool and optimizing a confidence-weighted ranking-fidelity reward, the distilled Qwen3-VL-8B raises mean \(\srcc\) from 0.504 to 0.709 across all three categories at single-pass inference cost, outperforming all open-source baselines and matching closed-source Gemini-3.1-Pro to within 0.04 \(\srcc\).

\paragraph{Limitations and future work.}
The matched dual-protocol design has been validated only on Chinese painting aesthetics, with \ppaint{} a 150-image diagnostic rather than a large-scale leaderboard. \ppdistill{} already transfers to oil painting and sketching on APDDv2 (Appendix~\ref{app:apddv2}), and extending it to non-art domains such as photographic quality, with more efficient bridges and validation across additional model families, is left to future work. We will release all expert annotations, the fused expert ground truth and the training corpus.

\newpage
\bibliographystyle{unsrtnat}
\bibliography{references}

\newpage
\appendix

\section{Dataset and Annotation Protocol Details}
\label{app:dataset}

\subsection{Dataset Composition}
\label{app:composition}

\ppaint{} consists of 150 expert-annotated paintings, with 50 paintings drawn from each of three categories (flower-and-bird, landscape, figure painting). Within each category, paintings are stratified into three quality tiers (good, medium, poor) based on a preliminary expert screening, and tier-internal selection covers a range of styles, periods, and techniques. Each painting receives pointwise ratings from five experts on five aesthetic dimensions, together with pairwise comparisons over 612 image pairs per category, also judged by five experts per pair. Aggregate annotation statistics are summarized in Table~\ref{tab:cpa_gt_stats}.

In addition to the 150-painting expert-annotated set, \ppaint{} includes a pool of approximately 9{,}000 unlabeled paintings (about 3{,}000 per category) used to construct the pseudo-labels for self-distillation in \S\ref{sec:vlm_diagnosis}. The released dataset comprises (i) the 150 paintings with their fused expert ground-truth scores on the 1--5 scale produced by anchored Elo and the anchored \dbtl{} cross-check, (ii) all raw expert pointwise scores and pairwise judgments, (iii) the majority-with-tie pairwise labels, (iv) the unlabeled image pool used for pseudo-label construction, and (v) the construction and analysis code. The full dataset and code will be released together with the paper.

\begin{table}[h]
\centering
\caption{\ppaint{} annotation statistics.}
\label{tab:cpa_gt_stats}
\scriptsize
\setlength{\tabcolsep}{4.8pt}
\renewcommand{\arraystretch}{1.08}
\begin{tabularx}{0.72\linewidth}{@{}>{\raggedright\arraybackslash}X>{\raggedleft\arraybackslash}p{0.24\linewidth}@{}}
\toprule
\textbf{Quantity} & \textbf{Value} \\
\midrule
Expert-annotated paintings & 150 \\
Painting categories & 3 \\
Aesthetic dimensions per painting & 5 \\
Experts per category / total & 5 / 15 \\
Pointwise raw scores & 3{,}750 \\
Pairwise image pairs annotated & 1{,}836 \\
Pairwise expert judgments & 45{,}900 \\
Unlabeled image pool for pseudo-labels & ${\sim}9{,}000$ \\
\bottomrule
\end{tabularx}
\end{table}

\subsection{Annotation Protocol}
\label{app:protocol}

\paragraph{Annotators.}
Each painting category is annotated by five domain experts, yielding 15 annotators in total. All annotators are research specialists in traditional Chinese painting affiliated with fine-arts academies, with extensive research or practical painting experience. Each annotator panel is restricted to its assigned category to ensure judgments reflect genuine domain familiarity.

\begin{figure}[h]
\centering
\includegraphics[width=\linewidth]{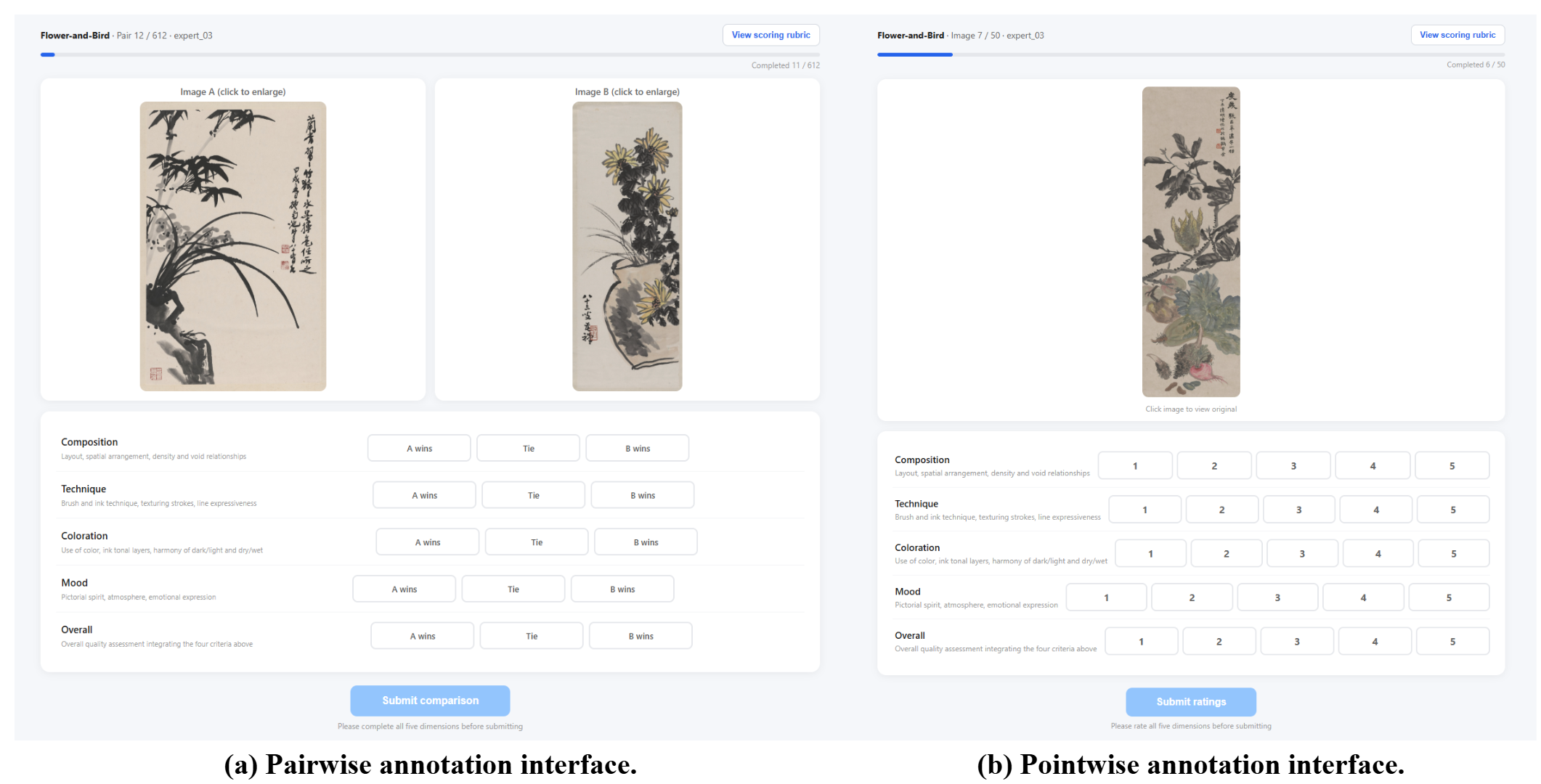}
\caption{Annotation interfaces used to collect \ppaint{}. \textbf{(a)} The pairwise interface presents two paintings from the same category and elicits a three-way choice (A wins, Tie, B wins) on each dimension. \textbf{(b)} The pointwise interface presents one painting and elicits an integer score from 1 to 5 on each of the five aesthetic dimensions. Both interfaces enforce a minimum viewing time and provide category-specific scoring rubrics through the side panel.}
\label{fig:interface}
\end{figure}

\paragraph{Interface.}
Both annotation tools are implemented as Flask web applications, illustrated in Figure~\ref{fig:interface}. The pairwise interface (Figure~\ref{fig:interface}a) displays two paintings from the same category side by side, and the annotator selects one of three options per dimension, A wins, Tie, or B wins. The pointwise interface (Figure~\ref{fig:interface}b) displays one painting and asks the annotator to assign an integer score from 1 to 5 on each of the five aesthetic dimensions, with dimension descriptions shown alongside the buttons. Both tools enforce a minimum viewing time before submission is allowed (5 seconds per image for pointwise and 10 seconds per pair for pairwise), and require the annotator to read the scoring rubric before the first task (10-second minimum stay on the guidelines page).

\paragraph{Pointwise score-scale calibration.}
Pointwise rating requires each annotator to internalize an absolute 1--5 scale, while pairwise comparison only needs relative judgments. To equalize the scale information available across the two protocols, each annotator previews a set of reference paintings before the pointwise phase begins. These references anchor the 1--5 score levels for every aesthetic dimension on the assigned category and are drawn from outside the 150-image dataset to avoid contamination. They remain accessible through a side panel throughout the pointwise session, allowing annotators to revisit the anchor levels at any time. The preview removes the scale-internalization burden that is intrinsic to pointwise rating but absent from pairwise comparison, so any remaining protocol gap can be attributed more to intrinsic protocol differences than to asymmetric scale guidance.

\paragraph{Workflow and quality control.}
Annotation proceeds in two phases separated by a two-day rest period. The pointwise phase is completed first for all images, followed by the pairwise phase on the same image set. This ordering prevents pairwise outcomes from biasing absolute score assignment, and the rest period reduces the chance that specific score values assigned during the pointwise phase are recalled and carried into the pairwise judgments. A residual familiarity effect with the image set cannot be fully ruled out and is treated as a limitation of the matched protocol. Three quantitative quality-control mechanisms are applied throughout. First, repeat annotations are inserted to measure intra-annotator consistency (10 repeated images per annotator with a minimum gap of 5 intervening items in the pointwise phase, and 30 repeated pairs per annotator with a minimum gap of 10 in the pairwise phase). Repeat-pair agreement rates range from 78.3\% to 93.3\% across annotators. Second, transitivity is verified for each annotator's pairwise judgments. The fraction of triples violating transitivity ranges from 0.10\% to 4.47\%, well below the chance level. Third, annotators whose rankings deviate most strongly from the group consensus are flagged for review. After the review, no annotator is excluded from the final dataset.

\paragraph{Annotation effort.}
Each judgment is logged with a millisecond-resolution timestamp. The per-annotator median time is 12 to 46 seconds per painting for pointwise rating and 21 to 28 seconds per pair for pairwise comparison. Aggregated across all 15 annotators, the total annotation effort amounts to approximately 115 expert-hours, of which roughly 7 hours are spent on pointwise rating and 108 hours on pairwise comparison.

\paragraph{Annotator consent and compensation.}
Annotators participated under an agreed expert-annotation arrangement and were informed that their judgments would be used for research and dataset construction. Compensation followed the agreed professional annotation rate, and the released dataset does not contain annotator identities or other personal information.

\subsection{Pairwise Annotation Budget}
\label{app:pair_budget}

A complete pairwise protocol over the 50 paintings in each category requires $\binom{50}{2}=1{,}225$ comparisons. \ppaint{} adopts a budget of 612 pairs per category, approximately half of the complete set, to retain dense local preference information while keeping expert workload manageable. This subsection provides empirical justification for the choice and characterizes the resulting comparison set.

\begin{figure}[h]
\centering
\includegraphics[width=0.92\linewidth]{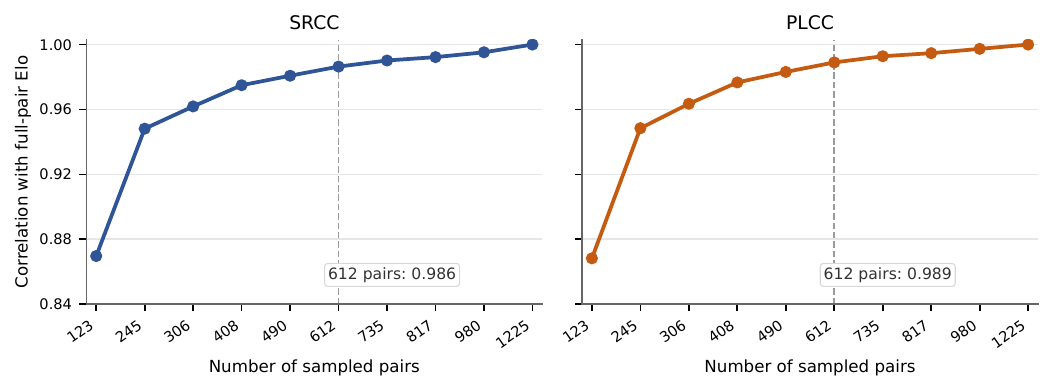}
\caption{Pairwise annotation budget pilot. Using Qwen3-VL-8B pairwise judgments as a proxy, we subsample different fractions of the complete 1{,}225-pair graph, run Elo aggregation, and compare the resulting scores with the full-graph Elo reference. Curves report means over category-dimension groups and random subsampling seeds. The dashed vertical line marks the 612-pair budget used in \ppaint{}.}
\label{fig:pair_budget}
\end{figure}

As a budget validation check, we test whether this reduced budget retains sufficient local information for Elo ranking, using pairwise judgments from Qwen3-VL-8B as a proxy. For each category-dimension group, we first aggregate the full 1{,}225 pairwise judgments into a reference Elo score. We then sample comparison graphs at fractions ranging from 10\% to 80\% of the complete graph, re-run Elo aggregation on each subsample, and compare the resulting scores with the full-graph reference using \(\srcc\) and \(\plcc\). As shown in Figure~\ref{fig:pair_budget}, the curves saturate well before the budget point. At 612 pairs the subsampled Elo scores reach 0.986 \(\srcc\) and 0.989 \(\plcc\) on average across category-dimension groups and random subsampling seeds, indicating that the chosen budget recovers nearly the full-graph ranking signal at roughly half the annotation cost.

We further characterize the connectivity of the 612-pair comparison graph by its diameter, defined as the longest shortest path between any two images. Across all three categories, the graph has diameter 2 and an average shortest path length of 1.47. Any two images are therefore connected by at most two intermediate comparisons, which limits the propagation of error along long transitive chains during Elo aggregation.

\subsection{Aesthetic Dimensions and Scoring Rubric}
\label{app:aesthetic_dimensions}

\ppaint{} evaluates each painting along five dimensions, namely brush technique (\emph{jifa}), coloration (\emph{shese}), composition (\emph{goutu}), conception (\emph{yijing}), and overall quality. The five-dimension design is motivated by three considerations.

First, decomposing aesthetic evaluation into explicit dimensions reduces the ambiguity of a single holistic score. When annotators are asked for one overall rating, different experts implicitly weight different internal criteria, with one emphasizing technique while another emphasizes mood, which produces apparent disagreement that reflects weighting differences rather than genuine quality disagreement. Five explicit dimensions make each criterion concrete and remove this confound.

Second, multi-dimensional annotation provides a fair comparison between the two protocols. With five structured dimensions, pointwise annotators receive the same level of guidance as pairwise annotators. Any remaining advantage of pairwise comparison under this matched setting is therefore more attributable to the protocol itself than to the granularity of the rating task.

Third, per-dimension analysis reveals where pairwise advantages are strongest. As reported in Appendix~\ref{app:dim_advantage}, the most subjective dimensions benefit most from pairwise comparison, a finding that a single-dimension design could not surface.

The five dimensions are grounded in traditional Chinese painting theory, and each is defined by a category-specific scoring rubric with five anchored levels. As an illustrative example, the highest level for coloration in flower-and-bird painting is described as exhibiting ``highly personal color use, whether the refreshing clarity of Yun Shouping or the vivid boldness of Qi Baishi, with color and ink fused in exceptional aesthetic harmony.''

\section{Protocol Diagnostics and Inter-Annotator Agreement}
\label{app:annotation_diagnostics}

This section provides extended evidence for the matched-protocol findings reported in \S\ref{sec:cpa_gt_protocol_findings}, organized into per-dimension diagnostics, annotator-level calibration, per-judgment agreement, and stability under annotator subsampling.

\subsection{Aggregate and Per-Dimension Diagnostics}
\label{app:protocol_viz}
\label{app:dim_advantage}

\begin{figure}[h]
\centering
\includegraphics[width=\linewidth]{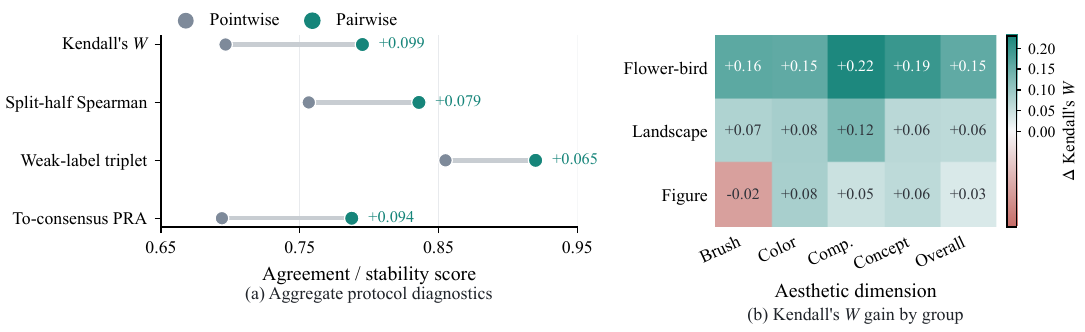}
\caption{Protocol effects under matched annotation. \textbf{(a)} Pairwise judgments exceed pointwise scores on all four aggregate diagnostics reported in Table~\ref{tab:protocol_diagnostics}. \textbf{(b)} The \kendallw{} advantage is broadly distributed, with 14 of 15 category-dimension groups showing a positive pairwise gain.}
\label{fig:protocol_diagnostics}
\end{figure}

This subsection complements Table~\ref{tab:protocol_diagnostics} (main paper) with two views, an aggregate visualization across categories and groups (Figure~\ref{fig:protocol_diagnostics}) and a per-dimension breakdown (Table~\ref{tab:dim_gap}).

Table~\ref{tab:dim_gap} reports the mean pairwise-minus-pointwise gap in \kendallw{} for each dimension, averaged across categories and computed against a per-image majority-with-tie reference derived from the five expert ratings. The pairwise advantage is positive for all five dimensions and for all 15 category-dimension groups. The magnitude of the gap varies, with overall and conception showing the largest values and coloration the smallest. The resulting ordering, $\textit{overall} > \textit{yijing} > \textit{goutu} > \textit{jifa} > \textit{shese}$, is consistent with the intuition that dimensions with greater rubric ambiguity benefit more from relative comparison than from absolute scoring.

\begin{table}[h]
\centering
\caption{Mean pairwise-minus-pointwise \kendallw{} gap by dimension, computed against a per-image majority-with-tie reference derived from the five expert ratings and averaged over the three painting categories. The gap is positive for every dimension across all 15 category-dimension groups.}
\label{tab:dim_gap}
\scriptsize
\setlength{\tabcolsep}{6pt}
\renewcommand{\arraystretch}{1.08}
\begin{tabular}{@{}lcc@{}}
\toprule
\textbf{Dimension} & \textbf{Mean gap} & \textbf{Median gap} \\
\midrule
Overall       & 0.275 & 0.221 \\
Yijing        & 0.258 & 0.260 \\
Goutu         & 0.206 & 0.181 \\
Jifa          & 0.194 & 0.149 \\
Shese         & 0.181 & 0.175 \\
\bottomrule
\end{tabular}
\end{table}

\subsection{Inter-Annotator Agreement and Calibration}
\label{app:annotator_agreement}

We characterize annotator-level agreement under both protocols using three complementary measures: Spearman and Kendall correlation between annotator-induced rankings, the mean rank displacement between each annotator and the group consensus, and the dispersion of these quantities across annotators. Pairwise annotation produces higher inter-annotator similarity, lower mean rank displacement, and tighter dispersion than pointwise annotation, as summarized in Table~\ref{tab:annotator_agreement}.

The annotator-to-consensus Spearman $\rho$ ranges from 0.777 to 0.923 under pairwise (spread of 0.146) versus 0.714 to 0.870 under pointwise (spread of 0.156). The standard deviation of the per-annotator rank error is also smaller under pairwise (0.99 versus 1.34). Beyond ranking-level agreement, pointwise annotators exhibit substantial scoring-pattern heterogeneity. The mean assigned score ranges from 2.30 to 3.65 across annotators, and the fraction of extreme scores (1 or 5) ranges from 10\% to 81\%. Pairwise annotators show less variation, with A-win rates ranging from 52.8\% to 70.0\% and tie rates from 0\% to 25.7\%. Higher tie rates correlate with lower consensus agreement, suggesting that excessive ties obscure genuine quality distinctions.

\begin{table}[h]
\centering
\caption{Inter-annotator agreement and calibration under the two protocols. All quantities are averaged over 15 category-dimension groups, with ranges reported across the 5 annotators per panel.}
\label{tab:annotator_agreement}
\scriptsize
\setlength{\tabcolsep}{5pt}
\renewcommand{\arraystretch}{1.08}
\begin{tabular}{@{}lcc@{}}
\toprule
\textbf{Metric} & \textbf{Pairwise} & \textbf{Pointwise} \\
\midrule
Annotator-pair Spearman $\rho$              & 0.739 & 0.621 \\
Annotator-pair Kendall $\tau$               & 0.552 & 0.532 \\
Mean rank error to consensus (out of 50)    & 5.38  & 6.64  \\
Std of rank error across annotators         & 0.99  & 1.34  \\
Annotator-to-consensus $\rho$ range         & [0.777, 0.923] & [0.714, 0.870] \\
\bottomrule
\end{tabular}
\end{table}

\subsection{Per-Judgment Agreement}
\label{app:per_judgment}

The aggregate metrics in Table~\ref{tab:annotator_agreement} compare ranking-level agreement after each protocol has been reduced to a per-image ranking. To measure agreement at the level of individual annotation events, we report two complementary analyses. The first projects both protocols onto a common pairwise judgment space by converting each annotator's pointwise scores into induced pairwise outcomes on the same 612 pairs (Table~\ref{tab:per_judgment}). The second compares raw 5-annotator agreement in each protocol's native annotation unit, that is, per-image ratings on five levels versus per-pair preferences on three outcomes (Table~\ref{tab:native_agreement}).

In the induced-pair view, preferences achieve higher agreement than induced pointwise judgments across all four metrics, with \fleissk{} of 0.373 versus 0.306 (12 of 15 category-dimension groups positive). In the native-unit view, pairwise annotators reach unanimous agreement on 42.4\% of pairs whereas pointwise annotators assign identical scores on only 5.7\% of images. The label space differs between the two protocols (3 outcomes for preferences and 5 scores for ratings), so the native-unit numbers are reported as descriptive context rather than a strict comparison. The strict, label-space-matched comparison is the induced-pair view in Table~\ref{tab:per_judgment}.

\begin{table}[h]
\centering
\caption{Per-judgment agreement on the same 612 pairs (averages over 15 category-dimension groups). Pointwise scores are converted to induced pairwise outcomes (A wins / Tie / B wins) before computing agreement, placing both protocols in a common label space.}
\label{tab:per_judgment}
\scriptsize
\setlength{\tabcolsep}{5pt}
\renewcommand{\arraystretch}{1.08}
\begin{tabular}{@{}lccc@{}}
\toprule
\textbf{Metric} & \textbf{Preferences} & \textbf{Ratings (induced)} & \textbf{Relative gap} \\
\midrule
\fleissk{}             & 0.373 & 0.306 & $+$22.1\% \\
\kalpha{}              & 0.373 & 0.309 & $+$20.7\% \\
Unanimous (5/5)        & 42.4\% & 28.7\% & $+$47.7\% \\
$\geq$4/5 agreement    & 65.6\% & 52.7\% & $+$24.5\% \\
\bottomrule
\end{tabular}
\end{table}

\begin{table}[h]
\centering
\caption{Raw 5-annotator agreement in each protocol's native annotation unit. Preferences are evaluated per-pair (3 labels: A wins / Tie / B wins) and ratings are evaluated per-image (5 labels: scores 1--5). Native-unit numbers are descriptive only because the label spaces differ. The strict comparison is reported in Table~\ref{tab:per_judgment}.}
\label{tab:native_agreement}
\scriptsize
\setlength{\tabcolsep}{5pt}
\renewcommand{\arraystretch}{1.08}
\begin{tabular}{@{}lcccc@{}}
\toprule
\textbf{Protocol} & \textbf{Unit} & \textbf{Labels} & \textbf{Unanimous (5/5)} & \textbf{$\geq$4/5} \\
\midrule
Preferences & per-pair  & 3 & 42.4\% & 65.6\% \\
Ratings     & per-image & 5 &  5.7\% & 23.1\% \\
\bottomrule
\end{tabular}
\end{table}

\subsection{Stability Under Annotator Subsampling}
\label{app:annotator_stability}

To evaluate how robust each protocol is to the size of the annotator panel, we conduct two subsampling analyses. The first removes one annotator at a time and measures the Spearman correlation between the leave-one-out ranking and the full-panel ranking. The second varies the subsample size $k \in \{1, 3, 5\}$ and computes the mean Spearman correlation between subsample rankings and the full-panel ranking, averaged over all category-dimension groups and all $\binom{5}{k}$ subsets.

Table~\ref{tab:annotator_stability} reports the results. Pairwise rankings are more robust to single-annotator removal, with mean Spearman $\rho = 0.975$ versus $0.953$ for pointwise and a higher Top-10 overlap (0.856 versus 0.707). Single-annotator pairwise rankings already reach $\rho = 0.860$, comparable to a 3-annotator pointwise panel ($\rho = 0.910$), indicating that pairwise comparison extracts more stable signal per annotator.

\begin{table}[h]
\centering
\caption{Stability of induced rankings under annotator subsampling. Leave-one-out values are averaged over 15 category-dimension groups and 5 leave-out trials each. Subsample values are averaged over all $\binom{5}{k}$ subsets for $k\in\{1,3\}$. The $k{=}5$ case is omitted because it trivially reproduces the full-panel reference for pointwise (deterministic averaging) and reaches $\rho{=}0.991$ for pairwise (Elo aggregation has minor non-determinism).}
\label{tab:annotator_stability}
\scriptsize
\setlength{\tabcolsep}{5pt}
\renewcommand{\arraystretch}{1.08}
\begin{tabular}{@{}lcc@{}}
\toprule
\textbf{Stability metric} & \textbf{Pairwise} & \textbf{Pointwise} \\
\midrule
Leave-one-out Spearman $\rho$              & 0.975 & 0.953 \\
Leave-one-out Top-10 overlap               & 0.856 & 0.707 \\
1-annotator subsample, mean $\rho$         & 0.860 & 0.791 \\
3-annotator subsample, mean $\rho$         & 0.947 & 0.910 \\
\bottomrule
\end{tabular}
\end{table}

\section{Fused Expert Ground-Truth Construction Details}
\label{app:gt_construction}

\subsection{Anchored Elo Construction}
\label{app:anchored_elo}
\label{app:anchors}

The canonical fused expert ground-truth construction estimates latent qualities through iterative Elo updates~\citep{elo1978rating}. For each observed pair $(i,j)$, the expected score is
\begin{equation}
E_{ij} = \sigma\!\bigl((q_i - q_j)/\tau\bigr), \label{eq:elo_expected}
\end{equation}
where $\tau$ is a temperature parameter. After observing outcome $o_{ij} \in \{1,\, 0.5,\, 0\}$ (win, tie, loss for image~$i$), the ratings are updated as
\begin{align}
q_i &\leftarrow q_i + K\,(o_{ij} - E_{ij}), \label{eq:elo_update_i}\\
q_j &\leftarrow q_j + K\,(1 - o_{ij} - E_{ji}), \label{eq:elo_update_j}
\end{align}
with step size $K$. After each full pass through all pairs, anchor images are pulled back toward their latent-space targets:
\begin{equation}
q_i \leftarrow q_i + \alpha\,(\bar{q}_i - q_i), \quad \forall\, i \in \mathcal{A}, \label{eq:elo_anchor}
\end{equation}
where $\alpha$ controls the anchor strength and $\bar{q}_i$ is the level-mapped Elo target corresponding to $\bar{s}_i$. This interleaving of Elo updates and anchor pull-back grounds the latent scale while letting pairwise data determine the ranking. The same sigmoid calibration (Eq.~\eqref{eq:sigmoid_cal}) maps the latent scores to the 1--5 aesthetic scale.

In the implementation, the step size $K$ takes a smaller value on pairs that involve at least one anchor image, which prevents anchor ratings from drifting too far from their targets during local updates. The step size also decays geometrically across passes to stabilize the final ratings. The complete set of hyperparameters is listed in Table~\ref{tab:elo_hp}.

\begin{table}[h]
\centering
\caption{Hyperparameters of the anchored Elo construction.}
\label{tab:elo_hp}
\scriptsize
\setlength{\tabcolsep}{6pt}
\renewcommand{\arraystretch}{1.08}
\begin{tabular}{@{}llc@{}}
\toprule
\textbf{Component} & \textbf{Hyperparameter} & \textbf{Value} \\
\midrule
Initialization & Initial rating $r_0$                                & 1500 \\
Pair update    & Step size on non-anchor pairs $K$                   & 32 \\
Pair update    & Step size on anchor-involving pairs $K_{a}$         & 6 \\
Pair update    & Per-pass step decay                                 & 0.995 \\
Anchor pull-back & Anchor strength $\alpha$                          & 0.15 \\
Anchor target  & Elo gap between adjacent anchor levels              & 400 \\
Optimization   & Number of passes                                    & 150 \\
Optimization   & Pair-shuffling random seed                          & 42 \\
\bottomrule
\end{tabular}
\end{table}

\paragraph{Anchor selection.}
For each category-dimension group, two anchor images are selected at each of the score levels 2, 3, and 4. Levels 1 and 5 are excluded to avoid boundary effects on the calibration target. Within each level, the candidates with the lowest inter-expert standard deviation are chosen. The total number of anchors per group ranges from 3 to 6 depending on data availability. Groups with fewer low-variance candidates at a given level (for example, figure painting composition) are harder to calibrate but still converge across the two anchored constructions.

\subsection{Anchored \dbtl{} Construction}
\label{app:anchored_btl}

As an independent construction, we model pairwise \textsc{A wins} / \textsc{Tie} / \textsc{B wins} outcomes with the Davidson extension of Bradley--Terry~\citep{davidson1970ties}. For a pair $(i,j)$,
\begin{align}
P(i \succ j) &= \frac{\exp(q_i)}{\exp(q_i)+\exp(q_j)+\nu\exp\!\bigl((q_i{+}q_j)/2\bigr)}, \label{eq:davidson_win}\\
P(i \sim j) &= \frac{\nu\exp\!\bigl((q_i{+}q_j)/2\bigr)}{\exp(q_i)+\exp(q_j)+\nu\exp\!\bigl((q_i{+}q_j)/2\bigr)}, \label{eq:davidson_tie}
\end{align}
with $P(j \succ i)$ defined symmetrically and $\nu$ capturing the propensity of experts to regard two paintings as indistinguishable. The latent qualities are estimated by maximizing the pairwise log-likelihood with an anchor regularizer:
\begin{equation}
\hat{\mathbf{q}} = \arg\max_{\mathbf{q}} \sum_{(i,j)} \log P(o_{ij} \mid q_i, q_j, \nu) \;-\; \lambda \sum_{i \in \mathcal{A}} (q_i - \bar{q}_i)^{2}, \label{eq:dbt_objective}
\end{equation}
where $o_{ij}$ is the observed outcome and $\lambda$ controls the strength of the anchor constraint. The subsequent sigmoid calibration (Eq.~\eqref{eq:sigmoid_cal}) maps the latent scores back to the 1--5 range. The tie parameter $\nu$ is estimated jointly with $\mathbf{q}$ during MAP optimization. Anchor targets are obtained by re-centering the corresponding pointwise medians, $\bar{q}_i = \bar{s}_i - 3$, so that the 1--5 score scale aligns with zero. We set $\lambda = 0.1$ and optimize the objective with L-BFGS-B for at most 2{,}000 iterations.

\subsection{Cross-Method Agreement}
\label{app:cross_method}

\begin{figure}[h]
\centering
\includegraphics[width=\linewidth]{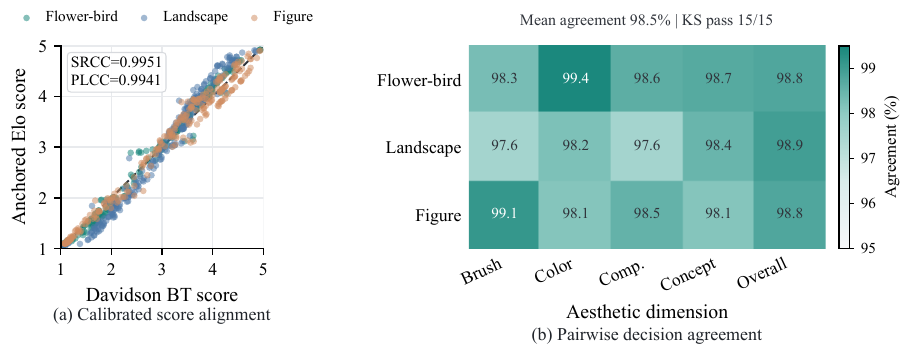}
\caption{Cross-method agreement of the fused expert ground truth. \textbf{(a)} Anchored Elo and anchored \dbtl{} produce nearly identical calibrated scores (\(\srcc=0.9951\), MAE\,$=0.1567$ on a 1--5 scale). \textbf{(b)} Their induced pairwise decisions agree at 98.5\% across category-dimension groups.}
\label{fig:fused_gt_reliability}
\end{figure}

This subsection reports cross-method agreement of the fused expert ground truth from three complementary views, an aggregate visualization (Figure~\ref{fig:fused_gt_reliability}), a four-method pairwise comparison (Table~\ref{tab:four_method}), and a per-group breakdown of the two anchored constructions (Table~\ref{tab:per_group_ks}).

Table~\ref{tab:four_method} reports the full pairwise agreement among four fused expert ground-truth construction variants, the two anchored methods used in this paper (Anchored Elo and Anchored \dbtl{}) together with their unanchored counterparts (Unanchored Elo and Unanchored \dbtl{}). The two anchored methods show the strongest agreement, with \(\srcc\) = 0.9951, MAE = 0.157, and 15/15 KS pass. The unanchored variants reach comparably high \(\srcc\) but exhibit substantially larger MAE (0.343--0.517) and lower KS pass rates (3/15--12/15), indicating that anchor regularization is essential for stable absolute-scale calibration even when ranking-level agreement is preserved.

\begin{table}[h]
\centering
\caption{Cross-method agreement for all six pairwise comparisons among four fused expert ground-truth construction variants. The four methods are Anchored Elo, Anchored \dbtl{}, Unanchored \dbtl{} (no anchor regularization), and Unanchored Elo (no anchor pull-back), giving $\binom{4}{2}=6$ method pairs. For each pair, SRCC and PLCC are computed between the two methods' calibrated 1--5 scores per category-dimension group, MAE is the mean absolute score difference on the same scale, Decision is the percentage of pairs where the two methods agree on the induced \textsc{A wins}/\textsc{Tie}/\textsc{B wins} outcome, and KS pass counts how many of the 15 category-dimension groups produce score distributions that pass the Kolmogorov--Smirnov test ($p > 0.05$). All values except KS pass are averaged over the 15 groups.}
\label{tab:four_method}
\scriptsize
\setlength{\tabcolsep}{4pt}
\renewcommand{\arraystretch}{1.08}
\begin{tabular}{@{}lccccc@{}}
\toprule
\textbf{Method pair} & \textbf{SRCC} & \textbf{PLCC} & \textbf{MAE} & \textbf{Decision} & \textbf{KS pass} \\
\midrule
Anchored Elo vs Anchored \dbtl{}        & 0.9951 & 0.9941 & 0.157 & 98.5\% & 15/15 \\
Anchored \dbtl{} vs Unanchored \dbtl{}  & 0.9970 & 0.9899 & 0.435 & 98.8\% & 5/15  \\
Anchored \dbtl{} vs Unanchored Elo      & 0.9967 & 0.9960 & 0.262 & 98.6\% & 12/15 \\
Anchored Elo vs Unanchored \dbtl{}      & 0.9897 & 0.9773 & 0.517 & 97.5\% & 3/15  \\
Anchored Elo vs Unanchored Elo          & 0.9896 & 0.9899 & 0.343 & 97.5\% & 12/15 \\
Unanchored \dbtl{} vs Unanchored Elo    & 0.9993 & 0.9918 & 0.282 & 99.5\% & 8/15  \\
\bottomrule
\end{tabular}
\end{table}

Table~\ref{tab:per_group_ks} reports the cross-method agreement between Anchored Elo and Anchored \dbtl{} for each of the 15 category-dimension groups. All groups pass the KS test ($p > 0.05$), with \(\srcc\) $\geq$ 0.989 and decision agreement $\geq$ 97.6\%, confirming that the close agreement summarized in Figure~\ref{fig:fused_gt_reliability} holds uniformly across categories and dimensions rather than being driven by a small number of well-behaved groups.

\begin{table}[h]
\centering
\caption{Per-group cross-method agreement between Anchored Elo and Anchored \dbtl{}. All 15 groups pass the KS test.}
\label{tab:per_group_ks}
\scriptsize
\setlength{\tabcolsep}{3.5pt}
\renewcommand{\arraystretch}{1.06}
\begin{tabular}{@{}llccccc@{}}
\toprule
\textbf{Category} & \textbf{Dimension} & \textbf{SRCC} & \textbf{PLCC} & \textbf{MAE} & \textbf{KS $p$} & \textbf{Decision} \\
\midrule
HuaNiao  & Goutu   & 0.997 & 0.995 & 0.120 & 0.717 & 98.6\% \\
HuaNiao  & Jifa    & 0.996 & 0.996 & 0.081 & 0.869 & 98.3\% \\
HuaNiao  & Overall & 0.997 & 0.995 & 0.120 & 0.869 & 98.8\% \\
HuaNiao  & Shese   & 0.999 & 0.997 & 0.162 & 0.272 & 99.4\% \\
HuaNiao  & Yijing  & 0.997 & 0.996 & 0.098 & 0.967 & 98.7\% \\
\midrule
RenWu    & Goutu   & 0.994 & 0.997 & 0.185 & 0.272 & 98.5\% \\
RenWu    & Jifa    & 0.998 & 0.998 & 0.074 & 0.869 & 99.1\% \\
RenWu    & Overall & 0.995 & 0.996 & 0.119 & 0.717 & 98.8\% \\
RenWu    & Shese   & 0.989 & 0.995 & 0.133 & 0.717 & 98.1\% \\
RenWu    & Yijing  & 0.994 & 0.994 & 0.078 & 0.967 & 98.1\% \\
\midrule
ShanShui & Goutu   & 0.994 & 0.988 & 0.258 & 0.396 & 97.6\% \\
ShanShui & Jifa    & 0.989 & 0.991 & 0.206 & 0.112 & 97.6\% \\
ShanShui & Overall & 0.997 & 0.992 & 0.225 & 0.717 & 98.9\% \\
ShanShui & Shese   & 0.994 & 0.992 & 0.196 & 0.717 & 98.2\% \\
ShanShui & Yijing  & 0.996 & 0.989 & 0.294 & 0.396 & 98.4\% \\
\bottomrule
\end{tabular}
\end{table}

\subsection{Elo Gap Calibration Ablation}
\label{app:elo_ablation}

The anchor Elo gap controls the spread between adjacent anchor levels in the latent target space and therefore acts as a calibration hyperparameter. Table~\ref{tab:elo_gap} reports cross-method agreement between Anchored Elo and Anchored \dbtl{} as the gap is varied. Reducing the gap from 1200 to 400 steadily improves all metrics. The final configuration with global sigmoid calibration and a gap of 400 achieves 15/15 KS pass and is adopted throughout the paper.

\begin{table}[h]
\centering
\caption{Elo gap calibration ablation. Each row reports cross-method agreement between Anchored Elo (with the specified configuration) and Anchored \dbtl{}. Gap is the Elo rating distance between adjacent anchor levels in the latent target space used by the anchor pull-back regularization (Eq.~\eqref{eq:anchor_reg}), so smaller Gap packs the five anchor levels more tightly in Elo space. The bottom three rows additionally use a Global Sigmoid (GSIG) that fits a single calibration curve by pooling anchors across all 15 category-dimension groups, instead of the per-group sigmoid used in the top four rows. Column definitions follow Table~\ref{tab:four_method}.}
\label{tab:elo_gap}
\scriptsize
\setlength{\tabcolsep}{4.5pt}
\renewcommand{\arraystretch}{1.08}
\begin{tabular}{@{}lccccc@{}}
\toprule
\textbf{Config} & \textbf{SRCC} & \textbf{PLCC} & \textbf{MAE} & \textbf{Decision} & \textbf{KS pass} \\
\midrule
Gap 1200 (baseline) & 0.9852 & 0.9864 & 0.241 & 96.9\% & 12/15 \\
Gap 900             & 0.9885 & 0.9911 & 0.187 & 97.4\% & 13/15 \\
Gap 700             & 0.9917 & 0.9936 & 0.144 & 97.9\% & 13/15 \\
Gap 500             & 0.9943 & 0.9950 & 0.128 & 98.4\% & 13/15 \\
\midrule
GSIG + Gap 350      & 0.9950 & 0.9932 & 0.185 & 98.4\% & 15/15 \\
GSIG + Gap 400      & 0.9951 & 0.9941 & 0.157 & 98.5\% & 15/15 \\
GSIG + Gap 450      & 0.9947 & 0.9945 & 0.142 & 98.4\% & 15/15 \\
\bottomrule
\end{tabular}
\end{table}

\section{Self-Distillation Implementation and Extended Results}
\label{app:impl}

\subsection{Pseudo-Label Generation}
\label{app:pseudo_label_algo}

\begin{table}[h]
\centering
\caption{VLM inference cost for pseudo-label construction. Each VLM call returns judgments or scores for all five aesthetic dimensions. Inference cost is the total number of VLM forward calls needed to label $n$ training images, where $N$ is the size of the Elo reference pool and $K$ is the number of pool comparisons per non-pool image. Concrete counts use $N{=}50$, $K{=}10$, and $n{=}3{,}000$. The exhaustive pairwise row reports a reference cost rather than an executed configuration.}
\label{tab:pseudo_label_stats}
\scriptsize
\setlength{\tabcolsep}{6pt}
\renewcommand{\arraystretch}{1.10}
\begin{tabular}{@{}lc@{}}
\toprule
\textbf{Pseudo-label source} & \textbf{Total VLM forward calls} \\
\midrule
Majority vote (32 pairwise trials per image) & $32n = 96{,}000$ \\
\pwdistill{} (raw VLM scores)        & $n = 3{,}000$ \\
\ppdistill{} bridge (ours)           & $\binom{N}{2} + (n-N)K = 30{,}725$ \\
Exhaustive pairwise reference        & $\binom{n}{2} = 4{,}498{,}500$ \\
\bottomrule
\end{tabular}
\end{table}

The offline preference-to-score bridge described in \S\ref{sec:bridge} produces continuous pseudo-scores for each unlabeled training image on each of the five aesthetic dimensions. Compared with exhaustive pairwise comparison, the bridge yields per-image pseudo-scores at modest VLM inference cost. Table~\ref{tab:pseudo_label_stats} summarizes the offline VLM inference budget required by each pseudo-label source.

The \ppdistill{} bridge is about 146 times cheaper than exhaustive pairwise comparison while still using pairwise judgments to construct per-image pseudo-scores. Algorithm~\ref{alg:bridge} gives the complete procedure.

\begin{algorithm}[h]
\caption{Preference-to-Score Bridge for Pseudo-Label Generation}
\label{alg:bridge}
\begin{algorithmic}[1]
\REQUIRE Unlabeled training images $\mathcal{X}=\{x_1,\dots,x_n\}$, pretrained VLM $\mathcal{M}$, pool size $N$, comparisons per image $K$, aesthetic dimensions $D{=}5$
\ENSURE Pseudo-scores $\{\tilde{s}_i^{(d)}\}$ for all $x_i \in \mathcal{X}$ and $d \in \{1,\dots,D\}$

\STATE \textbf{Stage 1: Ranking Pool Construction}
\STATE Sample $N$ images uniformly from $\mathcal{X}$ to form pool $\mathcal{P}$
\FOR{each pair $(x_i, x_j)$ in $\mathcal{P}$, $i < j$}
    \STATE Query $\mathcal{M}$ for A wins / Tie / B wins judgments across all $D$ dimensions
\ENDFOR
\FOR{each dimension $d = 1, \dots, D$}
    \STATE Run Elo updates over all $\binom{N}{2}$ pool pairs to obtain pool ratings $\{q_p^{(d)}\}_{p \in \mathcal{P}}$
\ENDFOR
\STATE Freeze pool ratings

\STATE
\STATE \textbf{Stage 2: Pool-Based Pseudo-Scoring}
\FOR{each remaining image $x_i \in \mathcal{X} \setminus \mathcal{P}$}
    \STATE Sample $K$ pool references without replacement
    \FOR{each sampled pool reference $x_p$}
        \STATE Query $\mathcal{M}$ for A wins / Tie / B wins judgments across all $D$ dimensions
    \ENDFOR
    \FOR{each dimension $d = 1, \dots, D$}
        \STATE Estimate $q_i^{(d)}$ via Bayesian posterior inference using $K$ pairwise outcomes and pool ratings as prior
    \ENDFOR
\ENDFOR

\STATE
\STATE \textbf{Stage 3: Sigmoid Calibration}
\FOR{each dimension $d = 1, \dots, D$}
    \STATE Compute $q_{\min}^{(d)}, q_{\max}^{(d)}$ across all images
    \FOR{each image $x_i \in \mathcal{X}$}
        \STATE $\tilde{s}_i^{(d)} \leftarrow 1 + \frac{4}{1 + \exp\!\left(-\left(\frac{6\,(q_i^{(d)} - q_{\min}^{(d)})}{q_{\max}^{(d)} - q_{\min}^{(d)}} - 3\right)\right)}$
    \ENDFOR
\ENDFOR
\RETURN $\{\tilde{s}_i^{(d)}\}$
\end{algorithmic}
\end{algorithm}

\subsection{GRPO Training}
\label{app:grpo}

We adopt Group Relative Policy Optimization (GRPO)~\citep{shao2024deepseekmath} for the online distillation stage (\S\ref{sec:grpo_method}). For each image $x_i$ with prompt $c$, the policy $\pi_\theta$ generates $G$ candidate responses. Let $r_k(x_i)$ denote the reward for the $k$-th response (computed from $R_{\text{rank}}^{(i,k)}$ in Eq.~\ref{eq:weighted_reward} plus auxiliary format and range terms). The relative advantage is estimated by within-group normalization,
\begin{equation}
a_k(x_i) = \frac{r_k(x_i) - \mu(r(x_i))}{\sigma(r(x_i))},
\label{eq:grpo_advantage}
\end{equation}
where $\mu(\cdot)$ and $\sigma(\cdot)$ are the mean and standard deviation of the $G$ rewards for image $x_i$. The policy is updated via the clipped surrogate objective
\begin{equation}
\mathcal{L}(\theta) = \frac{1}{BG}\sum_{i=1}^{B}\sum_{k=1}^{G} \min\!\left(\rho_k \cdot a_k,\; \mathrm{clip}(\rho_k, 1{-}\epsilon, 1{+}\epsilon) \cdot a_k\right) - \beta\, D_{\mathrm{KL}}\!\left(\pi_\theta \,\|\, \pi_{\mathrm{ref}}\right),
\label{eq:grpo_objective}
\end{equation}
where $\rho_k = \pi_\theta(o_k \mid c, x_i) / \pi_{\mathrm{old}}(o_k \mid c, x_i)$ is the importance ratio, $\epsilon$ is the clipping threshold, and $\beta$ controls the strength of KL regularization against the frozen reference policy $\pi_{\mathrm{ref}}$. Training hyperparameters and the hardware configuration are listed in Table~\ref{tab:grpo_hp}.

\begin{table}[h]
\centering
\caption{Training configuration for the online GRPO stage.}
\label{tab:grpo_hp}
\scriptsize
\setlength{\tabcolsep}{6pt}
\renewcommand{\arraystretch}{1.10}
\begin{tabular}{@{}llc@{}}
\toprule
\textbf{Component} & \textbf{Setting} & \textbf{Value} \\
\midrule
Backbone       & Base policy                     & Qwen3-VL-8B-Instruct \\
Sampling       & Group size $G$                  & 6 \\
Optimization   & Optimizer                       & AdamW \\
Optimization   & Learning rate                   & $1{\times}10^{-6}$ \\
Optimization   & Effective batch size            & 64 images per step \\
Optimization   & Gradient clipping               & 1.0 \\
Optimization   & Training epochs                 & 10 \\
Generation     & Maximum completion length       & 512 tokens \\
Inference      & Decoding                        & greedy, single forward pass \\
Hardware       & Accelerators                    & 8 $\times$ NVIDIA A100 80GB \\
Hardware       & Distributed strategy            & DeepSpeed ZeRO-3 \\
\bottomrule
\end{tabular}
\end{table}

\subsection{Prompt Templates}
\label{app:prompt}

Both the offline pseudo-label construction (\S\ref{sec:bridge}) and the online GRPO training (\S\ref{sec:grpo_method}) share a single prompt template, with the expert role embedded in the first line of the prompt itself rather than a separate system message. The pointwise variant assigns a continuous score in $[1.00, 5.00]$ on each of the five aesthetic dimensions and returns the answer in a fixed JSON schema. The pairwise variant presents the two paintings as the first and second images and selects \textsc{A}, \textsc{B}, or \textsc{TIE} on each of the same five dimensions. Both variants require the model to first emit its reasoning inside \texttt{<think>} \texttt{</think>} tags and then the final answer inside \texttt{<answer>} \texttt{</answer>} tags, matching the parser used by the reward function (Appendix~\ref{app:grpo}).

\smallskip
\noindent\fbox{\parbox{0.96\linewidth}{\small\ttfamily
\textbf{Pointwise prompt.}\\[2pt]
You are an expert in traditional Chinese painting criticism and aesthetics.\\[2pt]
Evaluate this Chinese painting on five criteria. Each must be a float from 1.00 to 5.00, rounded to two decimal places (1.00 = poor, 3.00 = moderate, 5.00 = excellent).\\[2pt]
1) technique\quad 2) coloration\quad 3) composition\quad 4) mood\quad 5) overall\\[2pt]
First output the thinking process in $\langle$think$\rangle$ $\langle$/think$\rangle$ tags and then output the final answer as valid JSON in $\langle$answer$\rangle$ $\langle$/answer$\rangle$ tags:\\[2pt]
\{``technique'': $\langle$1.00--5.00$\rangle$, ``coloration'': $\langle$1.00--5.00$\rangle$, ``composition'': $\langle$1.00--5.00$\rangle$, ``mood'': $\langle$1.00--5.00$\rangle$, ``overall'': $\langle$1.00--5.00$\rangle$\}
}}
\smallskip

\smallskip
\noindent\fbox{\parbox{0.96\linewidth}{\small\ttfamily
\textbf{Pairwise prompt.}\\[2pt]
You are an expert in traditional Chinese painting criticism and aesthetics.\\[2pt]
Compare two Chinese paintings (A is the first image, B is the second image) on five criteria. For each criterion, output ``A'' if the first painting is stronger, ``B'' if the second is stronger, or ``TIE'' if genuinely equal.\\[2pt]
1) technique\quad 2) coloration\quad 3) composition\quad 4) mood\quad 5) overall\\[2pt]
First output the thinking process in $\langle$think$\rangle$ $\langle$/think$\rangle$ tags and then output the final answer as valid JSON in $\langle$answer$\rangle$ $\langle$/answer$\rangle$ tags:\\[2pt]
\{``technique'': ``A''/``B''/``TIE'', ``coloration'': ``A''/``B''/``TIE'', ``composition'': ``A''/``B''/``TIE'', ``mood'': ``A''/``B''/``TIE'', ``overall'': ``A''/``B''/``TIE''\}
}}
\smallskip

\subsection{Closed-Source Model Versions}
\label{app:closed_model_versions}

Table~\ref{tab:closed_model_versions} lists the closed-source model versions used in the main comparison.

\begin{table}[h]
\centering
\caption{Closed-source model versions used in the main comparison.}
\label{tab:closed_model_versions}
\small
\setlength{\tabcolsep}{5pt}
\begin{tabular}{ll}
\toprule
\textbf{Model} & \textbf{Version} \\
\midrule
Gemini-3.1-Pro & \texttt{gemini-3.1-pro-preview} \\
Gemini-3-Flash & \texttt{gemini-3-flash-preview} \\
Claude-Sonnet-4.6 & \texttt{default} \\
Doubao-Seed-1.6-Vision & \texttt{doubao-seed-1-6-vision-250815} \\
Qwen3-VL-Plus & \texttt{qwen3-vl-plus-2025-12-19} \\
Qwen3.6-Plus & \texttt{qwen3.6-plus-2026-04-02} \\
GPT-5.4-Mini & \texttt{gpt-5.4-mini-2026-03-17} \\
\bottomrule
\end{tabular}
\end{table}

\subsection{Cross-Category Training and Supervision Granularity}
\label{app:cross_category}

Table~\ref{tab:cross_category} reports the full cross-category training comparison between \ppdistill{} and EvoQuality-style majority-vote distillation~\citep{chen2025evoquality}. Each method is trained on a single category and evaluated zero-shot on all three.

\begin{table}[h]
\centering
\caption{Effect of training data category. \ppdistill{} (ours) versus EvoQuality-style majority-vote distillation and \pwdistill{} (pointwise pseudo-score distillation). The best checkpoint per configuration is selected by mean \(\srcc\). Top two per column are \textbf{bolded} and \underline{underlined}.}
\label{tab:cross_category}
\scriptsize
\setlength{\tabcolsep}{2.8pt}
\renewcommand{\arraystretch}{1.10}
\begin{tabular}{@{}ll*{4}{c}*{4}{c}@{}}
\toprule
& & \multicolumn{4}{c}{\(\srcc \uparrow\)} & \multicolumn{4}{c}{\(\plcc \uparrow\)} \\
\cmidrule(lr){3-6} \cmidrule(lr){7-10}
\textbf{Method} & \textbf{Train} & \textbf{HN} & \textbf{RW} & \textbf{SS} & \textbf{Mean} & \textbf{HN} & \textbf{RW} & \textbf{SS} & \textbf{Mean} \\
\midrule
\pwdistill{}     & HN & 0.523 & 0.689 & 0.459 & 0.557 & 0.526 & 0.683 & 0.459 & 0.556 \\
                  & RW & 0.601 & 0.698 & 0.455 & 0.585 & 0.587 & 0.685 & 0.500 & 0.591 \\
                  & SS & 0.582 & 0.700 & 0.550 & 0.610 & 0.569 & 0.680 & 0.537 & 0.595 \\
\midrule
EvoQuality        & HN & 0.644 & 0.692 & 0.549 & 0.628 & 0.643 & 0.680 & 0.542 & 0.622 \\
                  & RW & 0.606 & 0.629 & 0.570 & 0.602 & 0.625 & 0.625 & 0.548 & 0.599 \\
                  & SS & 0.663 & 0.707 & 0.587 & 0.652 & 0.664 & 0.667 & 0.598 & 0.643 \\
\midrule
\ppdistill{} (ours)  & HN & \textbf{0.727} & \underline{0.713} & 0.686 & \underline{0.709} & \textbf{0.743} & \underline{0.693} & \underline{0.692} & \underline{0.709} \\
                  & RW & 0.676 & \textbf{0.715} & \textbf{0.754} & \textbf{0.715} & 0.688 & \textbf{0.706} & \textbf{0.738} & \textbf{0.711} \\
                  & SS & \underline{0.720} & 0.667 & \underline{0.700} & 0.696 & \underline{0.718} & 0.622 & 0.677 & 0.672 \\
\bottomrule
\end{tabular}
\end{table}

\paragraph{Ranking-level versus pair-level supervision.}
The gap between Calibrated Elo distillation ($0.709$) and EvoQuality-style majority-vote distillation ($0.628$) corresponds to a difference in supervision granularity. The majority-vote approach retains only the discrete outcome of each pair, discarding vote margins and imposing no consistency across pairs. The Calibrated Elo pipeline aggregates all pairwise outcomes into a global Elo ranking and maps it to the continuous 1--5 scale, which yields transitive orderings and finer score resolution. The difference is most visible on landscape painting, where per-pair judgments are noisier (Calibrated Elo achieves $0.686$ versus $0.549$ for majority vote). Across categories, the Calibrated Elo variant has a cross-category \(\srcc\) standard deviation of $0.021$, compared with $0.073$ for majority vote and $0.119$ for \pwdistill{}, consistent with global aggregation producing more stable supervision than pair-level labels.

\paragraph{Cross-training-source stability.}
Table~\ref{tab:cross_category} further reveals that Calibrated Elo supervision is robust to the choice of training category. Mean \(\srcc\) ranges from $0.696$ to $0.715$ (spread of 2.7\% relative) across three training sources. EvoQuality-style majority vote shows a wider range of $0.602$ to $0.652$ (spread of 8.3\% relative), and \pwdistill{} pseudo-scores exhibit a comparable spread of $0.557$ to $0.610$ (9.5\% relative). The ordering among methods is consistent across all training sources, with Calibrated Elo outperforming majority vote and \pwdistill{} regardless of the category used for training.

\subsection{Pair Weighting Ablation}
\label{app:extended_analysis}

Using the same ranking-pool seed for both variants, confidence weighting improves mean \(\srcc\) by 7.4\% relative over the unweighted Calibrated Elo variant. The gain appears across all three categories and is largest on landscape painting (11.9\% relative, from $0.613$ to $0.686$), indicating that confidence weighting is most useful when pseudo-label comparisons contain more uncertain or noisy local decisions. Per-category results are reported in Table~\ref{tab:pair_weight}.

\begin{table}[h]
\centering
\caption{Effect of confidence weighting under Calibrated Elo pseudo-labels. Both variants use the same offline pseudo-scores, ranking-pool seed, and training configuration. The only difference is whether the confidence weight $w_{iz}^{(d)}$ (Eq.~\ref{eq:pair_weight}) is applied.}
\label{tab:pair_weight}
\scriptsize
\setlength{\tabcolsep}{2.5pt}
\renewcommand{\arraystretch}{1.10}
\begin{tabular}{@{}>{\ \centering\arraybackslash}p{1.8cm}*{4}{c}*{4}{c}@{}}
\toprule
& \multicolumn{4}{c}{\(\srcc \uparrow\)} & \multicolumn{4}{c}{\(\plcc \uparrow\)} \\
\cmidrule(lr){2-5} \cmidrule(lr){6-9}
\textbf{Pair wt.} & \textbf{HN} & \textbf{RW} & \textbf{SS} & \textbf{Mean} & \textbf{HN} & \textbf{RW} & \textbf{SS} & \textbf{Mean} \\
\midrule
\cmark & \textbf{0.727} & \textbf{0.713} & \textbf{0.686} & \textbf{0.709} & \textbf{0.743} & \textbf{0.693} & \textbf{0.692} & \textbf{0.709} \\
\xmark & 0.688 & 0.677 & 0.613 & 0.660 & 0.692 & 0.650 & 0.637 & 0.660 \\
\bottomrule
\end{tabular}
\end{table}

\subsection{Sensitivity to Score Anchors and Ranking-Pool Seed}
\label{app:anchor_sensitivity}

Table~\ref{tab:anchor_sensitivity} studies the number of score anchors and the ranking-pool sampling seed used when mapping Elo ratings to the 1--5 score scale. All variants use flower-and-bird training images and the same self-distillation configuration, differing only in the score-anchor count or in the seed used to sample the reference ranking pool.

\begin{table}[h]
\centering
\caption{Sensitivity to score-anchor count and ranking-pool seed in Calibrated Elo pseudo-label construction. The best checkpoint is selected by mean \(\srcc\). Best per column in bold.}
\label{tab:anchor_sensitivity}
\scriptsize
\setlength{\tabcolsep}{2.4pt}
\renewcommand{\arraystretch}{1.10}
\begin{tabular}{@{}l*{4}{c}*{4}{c}@{}}
\toprule
& \multicolumn{4}{c}{\(\srcc \uparrow\)} & \multicolumn{4}{c}{\(\plcc \uparrow\)} \\
\cmidrule(lr){2-5} \cmidrule(lr){6-9}
\textbf{Setting} & \textbf{HN} & \textbf{RW} & \textbf{SS} & \textbf{Mean} & \textbf{HN} & \textbf{RW} & \textbf{SS} & \textbf{Mean} \\
\midrule
5 anchors, seed 11  & 0.710 & \textbf{0.715} & 0.668 & 0.698 & 0.602 & 0.561 & 0.531 & 0.565 \\
10 anchors, seed 11 & 0.726 & 0.683 & 0.680 & 0.696 & 0.720 & 0.678 & 0.688 & 0.695 \\
10 anchors, seed 42 & 0.727 & 0.713 & 0.686 & \textbf{0.709} & 0.743 & \textbf{0.693} & \textbf{0.692} & \textbf{0.709} \\
15 anchors, seed 11 & \textbf{0.729} & 0.678 & \textbf{0.702} & 0.703 & \textbf{0.765} & 0.634 & 0.686 & 0.695 \\
\bottomrule
\end{tabular}
\end{table}

Under the same ranking-pool sampling seed, reducing the number of anchors from 10 to 5 leaves mean \(\srcc\) close to the 10-anchor setting, but mean \(\plcc\) drops from $0.695$ to $0.565$. This indicates that a small anchor set can preserve coarse ordering while degrading alignment to the target score scale. The corresponding pseudo-labels concentrate many samples near the lower end of the score support, especially for composition, mood, and overall, reducing the resolution of the continuous supervision. Increasing the anchor count from 10 to 15 yields a small mean-\(\srcc\) change ($0.696$ to $0.703$) and nearly identical mean \(\plcc\), suggesting that the calibration benefit saturates around 10 anchors. We therefore use 10 anchors as the default setting in the main experiments.

\subsection{Robustness to Fused Expert Ground-Truth Construction Method}
\label{app:gt_robustness}

To test whether the distillation gain is an artifact of shared fused expert ground-truth construction methodology, we re-evaluate all self-distilled models against pure human pointwise targets, computed as the mean (or median) of the five expert pointwise scores per image-dimension pair. This pointwise target is constructed entirely from rating annotations and is therefore independent of the pseudo-label pipeline used in the offline stage (\S\ref{sec:bridge}).

Table~\ref{tab:gt_robustness} reports the results for all three self-distillation methods under the controlled flower-and-bird configuration used in the ablation tables. The method ordering is fully preserved under both the mean and median pointwise targets, with \ppdistill{} $>$ EvoQuality $>$ \pwdistill{} $>$ Base. \ppdistill{} achieves $0.621$ mean \(\srcc\) under the mean pointwise target and $0.594$ under the median pointwise target, corresponding to 34.7\% and 35.6\% relative improvements over the base model. Absolute correlations are lower than under the corresponding fused expert ground-truth evaluation ($0.709$), as expected, since the fused expert ground truth integrates both pairwise and pointwise signals and produces finer-grained rankings (\S\ref{sec:cpa_gt_fusion}). The preserved method ordering confirms that the distillation gain reflects genuine improvement in aesthetic ranking ability rather than an artifact of shared fused expert ground-truth construction methodology.

\begin{table}[h]
\centering
\caption{Evaluation against pure human pointwise targets (mean and median of five expert scores). All self-distillation models use the controlled flower-and-bird configuration from the ablation tables. $^\ddagger$ denotes baselines trained via the same Thurstone-model-based GRPO framework as \ppdistill{} with different pseudo-label sources. The method ordering is preserved under both pointwise targets.}
\label{tab:gt_robustness}
\scriptsize
\setlength{\tabcolsep}{2.2pt}
\renewcommand{\arraystretch}{1.10}
\begin{tabular}{@{}l*{4}{c}*{4}{c}@{}}
\toprule
& \multicolumn{4}{c}{\(\srcc \uparrow\)} & \multicolumn{4}{c}{\(\plcc \uparrow\)} \\
\cmidrule(lr){2-5} \cmidrule(lr){6-9}
\textbf{Method} & \textbf{HN} & \textbf{RW} & \textbf{SS} & \textbf{Mean} & \textbf{HN} & \textbf{RW} & \textbf{SS} & \textbf{Mean} \\
\midrule
\multicolumn{9}{l}{\textit{Target = human pointwise mean}} \\
Base (pretrained)         & 0.515 & 0.537 & 0.332 & 0.461 & 0.524 & 0.589 & 0.395 & 0.503 \\
\pwdistill{}$^\ddagger$  & 0.492 & 0.584 & 0.369 & 0.482 & 0.478 & 0.587 & 0.409 & 0.491 \\
EvoQuality$^\ddagger$     & 0.579 & 0.535 & 0.488 & 0.534 & 0.588 & 0.586 & 0.507 & 0.560 \\
\ppdistill{} (ours)          & \textbf{0.620} & \textbf{0.625} & \textbf{0.619} & \textbf{0.621} & \textbf{0.632} & \textbf{0.621} & \textbf{0.627} & \textbf{0.627} \\
\midrule
\multicolumn{9}{l}{\textit{Target = human pointwise median}} \\
Base (pretrained)         & 0.506 & 0.525 & 0.284 & 0.438 & 0.491 & 0.564 & 0.344 & 0.467 \\
\pwdistill{}$^\ddagger$  & 0.448 & 0.565 & 0.335 & 0.450 & 0.425 & 0.561 & 0.381 & 0.455 \\
EvoQuality$^\ddagger$     & 0.552 & 0.540 & 0.465 & 0.519 & 0.542 & 0.564 & 0.471 & 0.526 \\
\ppdistill{} (ours)          & \textbf{0.588} & \textbf{0.634} & \textbf{0.560} & \textbf{0.594} & \textbf{0.587} & \textbf{0.615} & \textbf{0.563} & \textbf{0.588} \\
\bottomrule
\end{tabular}
\end{table}

While Table~\ref{tab:gt_robustness} confirms that the ranking ordering of self-distilled models is preserved across evaluation-target choices, ranking metrics alone cannot detect score-scale collapse. To assess whether the distilled scores are also calibrated to the target score scale, Table~\ref{tab:gt_calibration} reports the mean absolute error (MAE), root mean squared error (RMSE), and the global predicted score mean for each method, evaluated against both the fused expert ground truth and the human pointwise mean target.

\begin{table}[h]
\centering
\caption{Absolute calibration of self-distilled models. MAE and RMSE are reported on a 1--5 scale against both the fused expert ground truth and the human pointwise mean. The rightmost two columns show the global predicted score mean across all 750 image-dimension pairs and its deviation from the fused expert ground-truth mean (2.95). The corresponding deviation from the human pointwise mean (3.06) is similar in magnitude. In the table headers, \textbf{Fused} refers to the fused expert ground truth and \textbf{Human} refers to the human pointwise mean target.}
\label{tab:gt_calibration}
\scriptsize
\setlength{\tabcolsep}{4.5pt}
\renewcommand{\arraystretch}{1.10}
\begin{tabular}{@{}lcccccc@{}}
\toprule
& \multicolumn{2}{c}{\textbf{MAE} $\downarrow$} & \multicolumn{2}{c}{\textbf{RMSE} $\downarrow$} & \multirow{2}{*}{\textbf{Pred.\ mean}} & \multirow{2}{*}{\textbf{$\Delta$ Fused}} \\
\cmidrule(lr){2-3} \cmidrule(lr){4-5}
\textbf{Method} & \textbf{Fused} & \textbf{Human} & \textbf{Fused} & \textbf{Human} & & \\
\midrule
Base (pretrained)         & 1.54 & 1.45 & 1.79 & 1.75 & 4.48 & $+1.53$ \\
\pwdistill{}$^\ddagger$  & 1.36 & 1.31 & 1.61 & 1.60 & 4.25 & $+1.30$ \\
EvoQuality$^\ddagger$     & 1.38 & 1.32 & 1.61 & 1.59 & 4.27 & $+1.32$ \\
\ppdistill{} (ours)          & \textbf{0.84} & \textbf{0.91} & \textbf{1.00} & \textbf{1.11} & \textbf{2.84} & $\mathbf{-0.11}$ \\
\bottomrule
\end{tabular}
\end{table}

The base model and both baselines exhibit severely mis-calibrated predictions despite their nontrivial ranking correlations. Their predicted score means range from $4.25$ to $4.48$, more than $1.2$ points above the fused expert ground-truth mean of $2.95$, and their MAE on the 1--5 scale ranges from $1.31$ to $1.54$. \ppdistill{} is the only method whose predicted mean ($2.84$) lies within $0.2$ points of either reference mean, and its MAE is roughly $40\%$ lower than the next-best baseline. The pattern provides a quantitative counterpart to the predicted-distribution collapse visualized in Figure~\ref{fig:firstfig}c.

\subsection{Per-Dimension Distillation Analysis}
\label{app:per_dim_distill}

Table~\ref{tab:per_dim_distill} breaks down the distillation results by aesthetic dimension (averaged over three categories). The improvement pattern mirrors the per-dimension pairwise advantage reported in \S\ref{app:dim_advantage}. The most subjective dimensions benefit most from Elo-based pseudo-labels. Mood shows the largest relative improvement over the base model for \ppdistill{} (130.4\%), followed by coloration (65.8\%), composition (46.4\%), and overall (45.8\%). In contrast, technique, where the base model already achieves $0.689$ \(\srcc\), improves by only 10.4\%. This alignment confirms that the pairwise advantage on subjective dimensions is not specific to human annotation but extends to VLM-generated pseudo-labels.

\begin{table}[h]
\centering
\caption{Per-dimension \(\srcc\) and \(\plcc\) (averaged over three categories, flower-and-bird training). All self-distillation methods use the same base model. Mood shows the largest gain, mirroring the per-dimension pairwise advantage in human annotation (Table~\ref{tab:dim_gap}).}
\label{tab:per_dim_distill}
\scriptsize
\setlength{\tabcolsep}{2.2pt}
\renewcommand{\arraystretch}{1.10}
\begin{tabular}{@{}l*{5}{c}c*{5}{c}c@{}}
\toprule
& \multicolumn{6}{c}{\(\srcc \uparrow\)} & \multicolumn{6}{c}{\(\plcc \uparrow\)} \\
\cmidrule(lr){2-7} \cmidrule(lr){8-13}
\textbf{Method} & \textbf{Tech.} & \textbf{Color.} & \textbf{Comp.} & \textbf{Mood} & \textbf{Over.} & \textbf{Mean} & \textbf{Tech.} & \textbf{Color.} & \textbf{Comp.} & \textbf{Mood} & \textbf{Over.} & \textbf{Mean} \\
\midrule
Base (pretrained)  & 0.689 & 0.351 & 0.509 & 0.310 & 0.508 & 0.473 & 0.706 & 0.432 & 0.548 & 0.299 & 0.578 & 0.513 \\
\pwdistill{}$^\ddagger$ & 0.704 & 0.515 & 0.627 & 0.277 & 0.660 & 0.557 & 0.721 & 0.475 & 0.603 & 0.328 & 0.654 & 0.556 \\
EvoQuality$^\ddagger$ & 0.710 & 0.577 & 0.575 & 0.580 & 0.698 & 0.628 & 0.736 & 0.544 & 0.595 & 0.549 & 0.685 & 0.622 \\
\ppdistill{} (ours)   & \textbf{0.761} & \textbf{0.582} & \textbf{0.745} & \textbf{0.714} & \textbf{0.741} & \textbf{0.709} & \textbf{0.763} & \textbf{0.573} & \textbf{0.721} & \textbf{0.720} & \textbf{0.769} & \textbf{0.709} \\
\bottomrule
\end{tabular}
\end{table}

\section{Cross-Dataset Generalization on APDDv2}
\label{app:apddv2}

\subsection{Setup}
\label{app:apddv2_setup}

APDDv2~\citep{lu2024apddv2} is a public aesthetic-assessment benchmark covering three painting categories with official human pointwise annotations. To test whether the calibrated-Elo supervision principle generalizes beyond Chinese painting, we apply \ppdistill{} and the EvoQuality baseline to two visually distinct subsets of APDDv2, oil painting and sketching. Oil painting is dominated by color and texture, whereas sketching is dominated by line and composition. Each subset provides 1{,}000 training images and 371 (oil) or 378 (sketching) test images. We confirm that the test sets contain no image duplicates with the \ppaint{} corpus by md5 hashing.

APDDv2 provides annotations over multiple aesthetic dimensions. For a controlled train--test evaluation, we use the three dimensions that are consistently available in both the training and test annotations: layout-and-composition, details-and-texture, and overall artistic quality. This yields a 3-dimensional 1--10 rubric, in contrast to the 5-dimensional 1--5 rubric used in \ppaint{}. Both methods inherit the configuration described in Appendix~\ref{app:grpo} with two adaptations. The reward functions are extended to operate over 3 dimensions, and both the score range in the prompt and the calibration mapping are extended to 1--10. All other hyperparameters remain unchanged.

\subsection{Cross-Domain Transfer}
\label{app:apddv2_transfer}

The two-domain APDDv2 setup also reveals an asymmetry in cross-domain transfer. Models trained on sketching transfer to oil painting more strongly than the reverse direction. Training \ppdistill{} on sketching yields a cross-domain \(\srcc\) of $0.690$ on oil painting, whereas training on oil painting yields only $0.587$ on sketching. A possible explanation is that sketching forces the policy to focus on layout and composition, which transfer well across genres, while oil-painting training admits richer surface cues such as color, texture, and brushwork that are more genre-specific. Per-domain in-domain versus cross-domain results are summarized in Table~\ref{tab:apddv2_transfer}.

\begin{table}[H]
\centering
\caption{Per-domain \(\srcc\) on APDDv2. Rows are grouped by training source. The diagonal entries report in-domain performance and the off-diagonal entries report cross-domain transfer. \textbf{Bold} marks the higher \(\srcc\) within each (training source, test domain) cell.}
\label{tab:apddv2_transfer}
\scriptsize
\setlength{\tabcolsep}{6pt}
\renewcommand{\arraystretch}{1.08}
\begin{tabular}{@{}llcc@{}}
\toprule
\textbf{Method} & \textbf{Train} & \textbf{Test = Oil} & \textbf{Test = Sketch} \\
\midrule
EvoQuality    & Oil  & 0.683 (in-domain) & 0.574 (cross) \\
\ppdistill{}  & Oil  & \textbf{0.752} (in-domain) & \textbf{0.587} (cross) \\
\midrule
EvoQuality    & Skt  & 0.620 (cross) & 0.560 (in-domain) \\
\ppdistill{}  & Skt  & \textbf{0.690} (cross) & 0.559 (in-domain) \\
\bottomrule
\end{tabular}
\end{table}

\end{document}